\pdfoutput=1
\RequirePackage{fix-cm}  %
\documentclass{article}
\PassOptionsToPackage{numbers,compress}{natbib}
\ifdefined\SubmissionCopy
  \usepackage[main]{neurips_2026}
\else
  \usepackage[preprint]{neurips_2026}
\fi
\ifdefined\SubmissionCopy\else
  \makeatletter\renewcommand{\@notice}{}\makeatother  %
\fi
\usepackage[T1]{fontenc}
\usepackage[utf8]{inputenc}
\usepackage{amsmath,amssymb,mathtools}
\usepackage{graphicx,booktabs,array,tabularx,wrapfig,needspace,multirow}
\usepackage[table]{xcolor}
\usepackage{microtype}
\usepackage[bottom,flushmargin]{footmisc}  %
\usepackage{hyperref}
\usepackage{acronym}
\usepackage{placeins}  %
\usepackage[colorinlistoftodos, textsize=small]{todonotes}

\newacro{CRAM}{Compact Reasoning Associative Memory}
\newacro{BPTT}{backpropagation through time}
\usepackage[capitalise,nameinlink]{cleveref}
\crefname{equation}{Eq.}{Eqs.}
\Crefname{equation}{Eq.}{Eqs.}
\crefname{section}{Sec.}{Secs.}
\Crefname{section}{Sec.}{Secs.}
\crefname{subsection}{Sec.}{Secs.}
\Crefname{subsection}{Sec.}{Secs.}
\crefname{subsubsection}{Sec.}{Secs.}
\Crefname{subsubsection}{Sec.}{Secs.}
\crefname{figure}{Fig.}{Figs.}
\Crefname{figure}{Fig.}{Figs.}
\crefname{table}{Table}{Tables}
\Crefname{table}{Table}{Tables}
\crefname{appendix}{App.}{Apps.}
\Crefname{appendix}{App.}{Apps.}
\hypersetup{colorlinks=true,linkcolor=black,citecolor=black,urlcolor=GivenColor}
\newif\ifmathcolors
\ifdefined\MonochromeMath\mathcolorsfalse\else\mathcolorstrue\fi
\definecolor{StateColor}{HTML}{2A78D6}
\definecolor{CouplingColor}{HTML}{EB6834}
\definecolor{GivenColor}{HTML}{4A3AA7}
\definecolor{TruthColor}{HTML}{18A372}
\definecolor{ReviewColor}{HTML}{B51D28}
\definecolor{AltPL}{HTML}{2A78D6}
\definecolor{AltMPF}{HTML}{1BAF7A}
\definecolor{AltCD}{HTML}{EB6834}
\definecolor{AltBPTTone}{HTML}{EDA100}
\definecolor{AltBPTT}{HTML}{6250D6}
\newcommand{\role}[2]{\ifmathcolors{\color{#1}#2}\else{#2}\fi}
\newcommand{\state}[1]{\role{StateColor}{#1}}
\newcommand{\coupling}[1]{\role{CouplingColor}{#1}}
\newcommand{\given}[1]{\role{GivenColor}{#1}}
\newcommand{\truth}[1]{\role{TruthColor}{#1}}

\newcommand{\note}[1]{\text{\scriptsize #1}}
\newcommand{\term}[2]{\underbrace{#1}_{\mathclap{\note{#2}}}}

\newcommand{\CRAM}{\textsc{CRAM}}
\newcommand{\BPTT}{BPTT}
\newcommand{\softmax}{\operatorname{softmax}}
\newcommand{\R}{\mathbb{R}}
\newcommand{\vect}[1]{\boldsymbol{#1}}

\newcommand{\logits}{\state{\vect v}}
\newcommand{\belief}{\state{\vect p}}
\newcommand{\puzzle}{\given{\vect x}}
\newcommand{\labels}{\truth{\vect y}}
\newcommand{\weights}{\coupling{\theta}}
\newcommand{\paircost}{\coupling{K}}
\newcommand{\symcost}{\coupling{W}}
\newcommand{\inputcost}{\coupling{B}}

\newcommand{\TableModelMark}{}
\newcommand{\TablePublished}[1]{%
  {\renewcommand{\TableModelMark}{\textsuperscript{\ensuremath{\dagger}}}#1}}
\newcommand{\TableScoreMark}{}
\newcommand{\TableRounded}[1]{%
  {\renewcommand{\TableScoreMark}{\rlap{\textsuperscript{*}}}#1}}
\newcommand{\TableSci}[2]{%
  \makebox[2em][r]{#1}\makebox[0.5em][c]{e}\makebox[1em][l]{#2}}
\newcommand{\TableScore}[2]{%
  \if\relax\detokenize{#2}\relax%
    \makebox[3.5em][r]{$#1$\TableScoreMark}\makebox[4em]{}%
  \else
    \makebox[3.5em][r]{$#1$\TableScoreMark}%
    \makebox[1.5em][c]{$\pm$}\makebox[2.5em][l]{$#2$}%
  \fi}
\newcommand{\TableTime}[2]{%
  \makebox[3.5em][r]{#1}\hspace{0.35em}\makebox[1.5em][l]{#2}}
\newcommand{\TableBest}[1]{{\bfseries\boldmath #1}}
\newcommand{\TableLabel}[1]{\textsf{#1}}
\newcommand{\TableCaptionLabel}[1]{{\small\sffamily #1}}
\newcommand{\TableHead}[1]{\multicolumn{1}{c}{\textbf{\TableLabel{#1}}}}

\newcommand{\TableBadge}[1]{%
  \leavevmode\raisebox{-0.35ex}{\includegraphics[height=2ex]{figures/icons/#1.pdf}}}
\newcommand{\TableBadgeSlot}[1]{%
  \makebox[2ex][c]{\if\relax\detokenize{#1}\relax\else\TableBadge{#1}\fi}}
\newcommand{\TableFamilies}[2]{%
  \leavevmode\TableBadgeSlot{#1}\hspace{2pt}\TableBadgeSlot{#2}\hspace{0.6em}}

\newcommand{\TableModel}[5]{%
  \TableFamilies{#1}{#2}%
  {\hypersetup{urlcolor=black}\href{#4}{#3}}\TableModelMark~%
  {\color{black!50}\hypersetup{citecolor=black!50}\citep{#5}}}
\newcommand{\TableCRAM}[1]{%
  \TableFamilies{graph}{loop}\textbf{\CRAM{}#1}}
\newcommand{\TableFRM}{\TableModel{}{loop}{FRM}{https://arxiv.org/abs/2606.29150}{frm2026}}
\newcommand{\TableTRM}{\TableModel{}{loop}{TRM}{https://arxiv.org/abs/2510.04871}{trm2025}}

\newcommand{\TableTRMPmarked}{\TableModel{}{loop}{TRMP\textsuperscript{\ensuremath{\S}}}{https://arxiv.org/abs/2610.18649}{trmp2026}}  %
\newcommand{\TableNCA}{\TableModel{}{loop}{NCA}{https://arxiv.org/abs/2609.36126}{etcheverry2026reasoning}}
\newcommand{\TableFPRM}{\TableModel{}{loop}{FPRM}{https://arxiv.org/abs/2606.18206}{fprm2026}}
\newcommand{\TableHRM}{\TableModel{}{loop}{HRM}{https://arxiv.org/abs/2506.21734}{hrm2025}}
\newcommand{\TableEqR}{\TableModel{}{loop}{EqR}{https://arxiv.org/abs/2605.21488}{eqr2026}}
\newcommand{\TableGRAM}{\TableModel{}{loop}{GRAM}{https://arxiv.org/abs/2605.19376}{gram2026}}
\newcommand{\TableEBR}{\TableModel{}{loop}{EBR}{https://energy-based-reasoners.github.io/}{ebr2026}}

\newcommand{\TableHand}{\TableFamilies{graph}{loop}\mbox{(hand-set)$^{\ddagger}$}}
\newcommand{\TableReference}[1]{\leavevmode{\color{black!45}#1}}
\newlength{\AltInitialWidth}
\newcommand{\AltInitial}[1]{\settowidth{\AltInitialWidth}{\textbf{#1}}\textbf{#1}\llap{\makebox[\AltInitialWidth][c]{{\color{black!45}\setlength{\fboxsep}{0.4pt}\setlength{\fboxrule}{0.7pt}\fbox{\phantom{\textbf{#1}}\vphantom{L}}}}}}
\newcommand{\AltClass}[1]{\kern1.1pt\AltInitial{#1}\kern1.1pt\hspace{0.4em}}
\newcommand{\KeepNote}{\postdisplaypenalty=10000\relax}
\newcommand{\EqNote}[1]{\par\nobreak\vspace{\dimexpr-1.1\baselineskip+0.15cm\relax}\nobreak{\small\interlinepenalty=10000\noindent{\color{black!55}#1}\par}\smallskip}

\makeatletter
\renewcommand{\paragraph}{%
  \@startsection{paragraph}{4}{\z@}%
                {0.5ex \@plus 0.3ex \@minus 0.1ex}%
                {-1em}%
                {\normalsize\bf}%
}
\makeatother
 \title{
  Classical Energies Learn to Reason 1,000$\times$ Faster than Looped Models
  }
\newcommand{\affil}[1]{\textsuperscript{\normalfont#1}}  %
\author{%
  Benjamin Hoover\affil{1} \qquad Hugh Blayney\affil{1} \qquad James C. R. Whittington\affil{1}
}
\begin{document}
\maketitle
\ifdefined\SubmissionCopy\else
  \begingroup
  \renewcommand\thefootnote{}
  \begin{NoHyper}%
  \footnotetext{\affil{1}Prefrontal, London, United Kingdom. Correspondence to Benjamin Hoover \texttt{<ben@prefrontal.ai>}, James C.R. Whittington \texttt{<james@prefrontal.ai>}}
  \end{NoHyper}
  \endgroup
\fi
\begin{abstract}
Looped models have emerged as a prominent paradigm for reasoning, with successes on simple reasoning puzzles such as Sudoku-Extreme and Maze-Unique. 
However, these models can be unstable, slow to train, and use many parameters and FLOPs to solve simple puzzles. 
Descent on an energy not only mitigates these issues but also offers advantages: built-in early stopping, compute that adapts to problem difficulty, fast training without differentiating through a solver, and interpretable weights that learn the constraints of the puzzle.
Here we demonstrate this with a learned version of Hopfield \& Tank's energy from the 80s, which we term \CRAM{}, solves 99.996\% of Sudoku-Extreme and 100\% of Maze-Unique test puzzles after 42\,ms and 132\,ms of training on one GPU, at least 1,000${\times}$ faster than looped models on these tasks.
\CRAM{} also matches baseline looped models at their highest accuracy on Sudoku-Extreme using at least 282${\times}$ fewer inference FLOPs.
Our work suggests two things. First, that energies are an extremely promising direction for training reasoning from data. Second, a caution against using trivially solvable benchmarks like Sudoku and Maze to test the scalability of looped reasoning techniques.
\end{abstract}

\section{Introduction}
\begin{figure}[t]
\centering
\includegraphics[width=\linewidth]{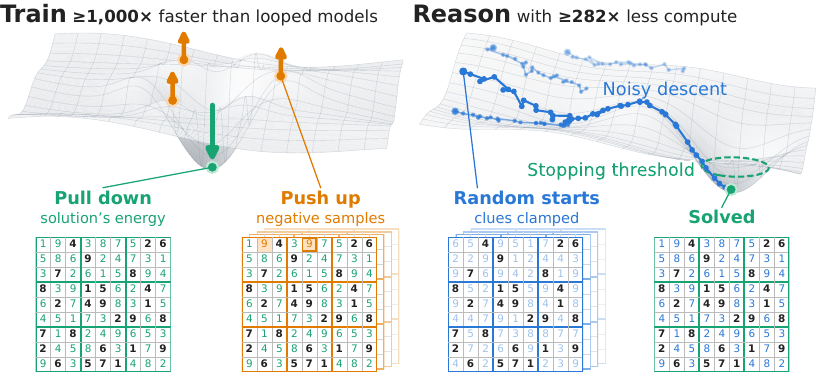}
\caption{\textbf{\CRAM{} learns an energy from solved puzzles in seconds, and the energy tells the search when to stop.}
\emph{Training} (left) pulls down the energy of each solution and pushes up the energy of negative samples, \emph{without} differentiating through a reasoning trajectory (shown is a pseudolikelihood objective, \cref{eq:pl}).
On one H200, training takes 42\,ms on Sudoku-Extreme and 132\,ms on Maze-Unique, at least 1,000${\times}$ faster than the looped models of \cref{tab:main}.
\emph{Reasoning} (right) clamps the clues of a new puzzle, starts several chains from random fills of the other cells (light-blue guesses; one board per chain), and minimizes the energy one cell at a time, using at least 282${\times}$ fewer inference FLOPs (\cref{fig:compute}) than looped models at matched accuracy.
The chains are independent, and the descent can be noisy, which lets a chain escape spurious local minima.
The search stops when any chain's energy falls below a stopping threshold fitted from the training solutions.
}
\label{fig:overview}
\end{figure}
An emerging new paradigm for reasoning is looping, in which a layer, or block of layers, is repeatedly applied to produce an answer. Small looped models \citep{hrm2025,trm2025} (millions of parameters) have had successes on simple benchmarks such as Sudoku and mazes. However, they are slow to train in part due to their use of truncated backprop-through-time (BPTT), and the reasons behind their successes remain unclear \citep{team2025hidden,ren2026your}. 

Energy minimization is a related framework that offers a principled approach to looping. Here, each loop step corresponds to a gradient descent step on an energy function. Minimizing energy offers many potential benefits, such as guaranteed stability, early stopping, 
and efficient training methods that bypass the need for BPTT.
Indeed, looped reasoners have begun adopting the ideas of energy-based networks without adopting an explicit energy. Some train their updates to converge to attractors or fixed points, a guarantee intrinsic to energy network dynamics \citep{eqr2026,fprm2026}. Others learn when to stop with a separate halting network \citep{act2016,pondernet2021,hrm2025,trm2025}.
Each of these desirable properties is generally added on top of the looped network through an extra head, loss, or module, and the network is trained by backpropagating through its repeated applications.
With looped models rebuilding the desirable properties of an energy one at a time, we ask: why not learn the energy directly?

It's already known that constraint puzzles like Sudoku have a corresponding energy that, when minimized, solves the puzzle \citep{hopfieldsudoku2008}. In the 1980's, Hopfield and Tank showed how to hand-design an energy function that solves certain classes of logic problems \citep{hopfieldtank1985}. It is, however, less well understood if an explicit energy can be learned from data to solve a wide array of reasoning problems.

Here, we develop a trainable version of the Hopfield-Tank energy and then learn the weights directly from puzzles. 
We call this tiny trainable energy, together with the recurrent dynamics that descend it at inference time, a Compact Reasoning Associative Memory (\CRAM{}).We call it this because, as in an Amari--Hopfield network \citep{amari1972,hopfield1982}, its weights \emph{memorize associations} between pairs of valid cells, and descent on the energy completes a corrupted pattern (i.e., a puzzle provided with only its clues).
\CRAM{}'s energy is a special case of the energies of modern Hopfield networks \citep{denseam2016,amtutorial2025}. 

In this work we: 1) show that \CRAM{} trains at least 1,000${\times}$ faster than modern looped models and reasons with at least 282${\times}$ fewer inference FLOPs on a variety of different puzzles (Sudoku-Extreme \citep{hrm2025}, Maze-Unique \citep{eqr2026}, and the N-Queens and graph-coloring sets of \citet{gram2026}); 2) show that \CRAM{} provides a stopping test that allows compute to scale with the difficulty of a puzzle;  3) show that \CRAM{} is data-efficient, learning interpretable weights from Sudoku-Extreme using an order of magnitude fewer training puzzles than looped models; 4) show that local training objectives train fastest compared to other methods of training the same energy --- significantly faster than BPTT; and 5) show that a hand-set \CRAM{} energy with few unique weights can solve every test puzzle from every benchmark exactly.  

\CRAM{} represents the state of the art on speed and efficiency of looping to solve constraint puzzles, by several orders of magnitude. 
However, since these puzzles can be solved easily with so few parameters using \CRAM{}, our work suggests that high accuracy on these benchmarks are not indicative of future performance on harder tasks.
Indeed, \CRAM{}'s simple energy on its own would struggle to learn more complex reasoning tasks like ARC-AGI~\cite{chollet2019}, though we do show how a simple latent extension of \CRAM{} can solve Maze-Hard near perfectly in \cref{sec:maze-hard-latents}. 
Nevertheless, the dramatic speed-up that this energy provides on the presented tasks offers a potential path forward for data-efficient general reasoning capabilities.

\section{Background}
\label{sec:background}

\CRAM{}'s energy is a form of \textbf{pairwise graphical model} in that it is a sum of costs on and between cells, as in the energy of a Markov random field (MRF) \citep{besag1974,wainwright2008}.
Learning these costs goes by several names, such as constraint programming \citep{schiex1995vcsp} or constraint acquisition \citep{bessiere2017conacq,modelseeker2012,deraedt2018}. Recent methods show that you can learn pairwise costs for Sudoku directly from solved puzzles, using approximate likelihood methods like those studied in this work \citep{brouard2020,defresne2023,epll2025}. 
However, to infer solutions to new puzzles, these methods pass the learned constraints to an external solver \citep{cooper2010softarc}.
\CRAM{} keeps this learning problem but uses neural optimization methods to replace the external solver with recurrent search on its own energy \citep{hopfieldtank1985,petersonsoderberg1989}.

\textbf{Energy-based networks} treat an energy over a large search space of possible solutions as the objective of both training and inference. 
Training shapes the energy to be low at observed data; inference or ``reasoning'' minimizes the energy to sample low-energy solutions (see \cref{fig:overview}) \citep{amari1972,hopfield1982,lecun2006energy}. 
Hopfield and Tank used a version of the Hopfield Network with continuous neurons to solve constraint problems like the Traveling Salesman Problem with hand-set weights \citep{hopfield1984,hopfieldtank1985}, and later work did the same for Sudoku \citep{hopfieldsudoku2008}.
\CRAM{} writes these energies in a mean-field free energy form with learnable weights \citep{petersonsoderberg1989}. 
This form approximates the free energy of a spin glass or Boltzmann machine \citep{edwards1975,sherrington1975,ackley1985}, which have similarly been applied to logic and combinatorial problems \citep{hinton1989dbm,danjou1993,tran2023lbm}.
However, none of these energies have been trained by observing puzzle solutions alone.

\textbf{Modern looped reasoners} use deep, weight-tied networks to learn puzzles directly from data.
HRM and TRM popularized this design for Sudoku, mazes and ARC, building on recurrent relational networks and universal transformers \citep{hrm2025,trm2025,chollet2019,rrn2018,universaltransformer2019}.
Later works varied the architecture but kept the same training approach: backpropagating through the reasoning trajectory of a depth-recurrent model to refine the model's prediction for a specified input \citep{urm2025,gram2026,ptrm2026,serrm2026,mgdm2025,adaptive2025,loopedflows2026,trmp2026}.
Some approaches use diffusion- and flow-style training to avoid BPTT \citep{ired2024,mgdm2025,adaptive2025,frm2026,fixedpointflows2026,loopedflows2026}.
Others engineer fixed point behavior into deep models, training the update loop to converge to stable equilibria \citep{eqr2026,fprm2026,frm2026,fixedpointflows2026,attractor2026}.
This last direction holds promise, since Hopfield \& Tank already showed that these reasoning puzzles are naturally suited to fixed-point problems.

\textbf{Energy-based looped models} encourage convergence by training an energy on top of a deep network, seeking to combine the best of looped models with advantages of Hopfield Networks and Boltzmann Machines \citep{irem2022,ired2024,ebt2025,akorn2025,neupsl2023,erm2026,ebr2026}.
For example, concurrent work Energy-Based Reasoners (EBR) \citep{ebr2026} showcases several advantages of energy for looped model reasoning: it trains an energy on the latent states of a TRM-style transformer, descends it with noisy projected gradient steps, and stops when the gradient norm is small \citep{ebr2026}. 
Though similar in spirit, these networks are much larger and slower than \CRAM{}, and all of them learn through BPTT.  %
\Needspace{18\baselineskip}
\section{Method}
\label{sec:method}
\label{sec:inference}
\label{sec:learning}

The reasoning tasks studied in this work can be represented via $M$ answer variables called \emph{cells}, each with up to $C$ candidate values. Therefore, an energy is assigned to each $M\times C$ belief array $\belief$, whose (non-negative) entry $\state{p}_{ia}$ gives the probability of candidate value $a$ at cell $i$.
Simply, $\belief_i$ is modeled as a softmax distribution over possible answers for cell $i$.

\CRAM{}'s energy can be written as
\begin{equation}
E_{\vect{\weights}}(\belief)
=\term{\beta^{-1}\sum_i\belief_i^{\!\top}\log\belief_i}{negative entropy}
+\term{\frac12\sum_{i\ne j}\belief_i^{\!\top}\vect{\paircost}_{ij}\belief_j}{interaction energy}
+\term{\sum_i\vect{\inputcost}_i^{\top}\belief_i}{input},
\label{eq:core}\KeepNote
\end{equation}
\EqNote{Throughout, \state{blue} identifies dynamic states and \coupling{orange} learned weights.}
with learnable parameters $\vect{\weights}$ and inverse temperature $\beta>0$. 
For each pair of cells $i,j$, the matrix $\vect{\paircost}_{ij}\in\R^{C\times C}$ contains trainable ``costs'' for all pairs of candidate values.
The vector $\vect{\inputcost}_i\in\R^C$ is a ``bias'' that describes the cost of each belief at cell $i$.
When $\belief_i$ is one-hot, the entropy term is zero, and \cref{eq:core} is simply the sum of the pair and input costs of the chosen values.
This energy is a mean-field free energy, which extends the energy of a continuous Hopfield network to $C$-state neurons~\citep{petersonsoderberg1989,hopfield1984}.

\paragraph{Reasoning as energy minimization.}
\CRAM{} infers solutions by recurrently minimizing the energy of \cref{eq:core} with a puzzle's givens held fixed; this is termed \emph{relaxation}.
Each cell has an unconstrained recurrent neuron state $\logits_i\in\R^C$ which is used as the logits for the belief $\belief_i = \softmax(\beta\logits_i)$ over possible answer states.
With time constant $\tau_i>0$, free neuron states evolve as the derivative of the energy,
\begin{equation}
\label{eq:flow}
\tau_i\frac{d\logits_i}{dt}
= -\logits_i-\sum_j\vect{\symcost}_{ij}\belief_j-\vect{\inputcost}_i
\simeq -\nabla_{\belief_i} E_{\vect{\weights}}(\belief),\KeepNote
\end{equation}
\EqNote{$\simeq$ denotes equality up to an additive scalar $c$, leaving $\belief_i = \softmax(\beta\logits_i + c) = \softmax(\beta\logits_i)$ unchanged.}
where $\vect{\symcost}_{ij}=(\vect{\paircost}_{ij}+\vect{\paircost}_{ji}^{\top})/2$ for $i\ne j$ pops out from the energy gradient, and $\vect{\symcost}_{ii}=\vect{0}$ (because the energy excludes self-interactions). We are able to update $\logits_i$ using the gradient of $E$ w.r.t. $\belief_i$ because the softmax is a monotonic function, so convergence to the same fixed $\belief_i^\star$ is still guaranteed. With equal time constants $\tau$ and step size $\eta=\Delta t/\tau$, an Euler step of \cref{eq:flow} is:
\begin{equation}
\label{eq:euler}
\logits_i\leftarrow(1-\eta)\logits_i-\eta\Big(\sum_j\vect{\symcost}_{ij}\belief_j+\vect{\inputcost}_i\Big).
\end{equation}
With $\eta=1$, this is the one-step update $\logits_i\leftarrow-\sum_j\vect{\symcost}_{ij}\belief_j-\vect{\inputcost}_i$, so each iteration is one layer of recurrence, as in looped models.
For fixed parameters and temperature, the deterministic descent of \cref{eq:flow} guarantees that the energy never increases in continuous time (i.e., $dE/dt\le0$, see \cref{app:descent}). We find that randomness during inference is important to help escape local minima and explore distinct solutions, so for our experiments we use a form of Gibbs sampling known as focused Gibbs search which prioritizes cell updates based on their local energy \citep{hopfieldtank1985,gemangeman1984}.
The energy also provides a natural early-stopping criterion: search can stop either when a board's energy is at most a threshold fitted from the training solutions, or when no single-cell change lowers the energy (\cref{app:recipes}).

\begin{table}[b]
\caption{\textbf{Supplied priors by task.}
$M{\times}C$ is the shape of the answer grid (num cells${\times}$num values).
NParams is the total number of parameters available to train.
NChains is the number of independent search chains per input: for N-Queens and graph coloring it is $N$, which varies in \cref{tab:nq-gc} ($N{=}1$ or $N{=}20$).
Init is the value $\state{x}_i$ at the start of a search chain; $\mathcal U(C)$ draws it uniformly from the $C$ values.
}
\label{tab:choices}
\fontsize{7.8}{9.4}\selectfont
\setlength{\tabcolsep}{3pt}
\renewcommand{\arraystretch}{1.08}
\begin{tabularx}{\linewidth}{@{}l@{\hspace{7pt}} >{\raggedright\arraybackslash\hsize=0.72\hsize}X >{\raggedright\arraybackslash\hsize=1.2\hsize}X >{\raggedright\arraybackslash\hsize=0.96\hsize}X >{\raggedright\arraybackslash\hsize=1.12\hsize}X@{}}
\toprule
 & {\fontsize{7.2}{9}\selectfont\textbf{\TableLabel{\mbox{Sudoku-Extreme}}}} & {\fontsize{7.2}{9}\selectfont\textbf{\TableLabel{Maze-Unique}}} & \mbox{{\fontsize{7.2}{9}\selectfont\textbf{\TableLabel{N-Queens}}} {\fontsize{6}{7}\selectfont\TableLabel{$8{\times}8$ ($10{\times}10$)}}} & \mbox{{\fontsize{7.2}{9}\selectfont\textbf{\TableLabel{Graph coloring}}} {\fontsize{6}{7}\selectfont\TableLabel{8 (10)}}}\\
\midrule
$M{\times} C$ & $81{\times}9$ & $900{\times}2$ & $64{\times}2$ ($100{\times}2$) & $8{\times}3$ ($10{\times}3$)\\
$\vect{\paircost}_{ij}$ & $9{\times}9$ per cell pair & $2{\times}2$ per offset in $3{\times}3$ window & one cost per offset & $3{\times}3$ per pair + $3{\times}3$ per edge\\
$\vect{\inputcost}_i$ & free per cell & shared by all cells & shared by all cells & free per cell (vertex)\\
\textbf{NParams} & 263K & 18 & 113 (181) & 7.3K (18.7K)\\
\textbf{Priors} & digit augmentation & locality, translation invariance & relative position only & none\\
\textbf{Input} & clues clamped & walls, endpoints clamped & given queens clamped & edges select per-edge tables\\
\textbf{NChains} & 4 & 1 & $N$ & $N$\\
\textbf{Init} & $\state{x}_i\sim\mathcal U(C)$ & $\state{x}_i{=}1$ (open cells on) & $\state{x}_i{=}0$ (no queen) & $\state{x}_i\sim\mathcal U(C)$\\
\bottomrule
\end{tabularx}
\end{table}
\begin{table}[!t]
\centering
\caption{\textbf{\CRAM{} trains faster and is more accurate and efficient than every looped baseline.}
The table reports accuracies and FLOPs at each model's listed inference setting.
Note that we also report a hand-set \CRAM{} energy (\TableReference{gray}).
\TableCaptionLabel{Solved}: mean {\color{black!50}$\pm$stdev} test accuracy (\%) over 3 training seeds.
\TableCaptionLabel{Params}: nonzero, contributing params of the evaluated model.
\TableCaptionLabel{FLOPs/puzl}: avg inference FLOPs per puzzle, including all search chains and failed searches (\cref{app:opcounts}).
\TableCaptionLabel{Fit time}: wall clock of training seed 0 on one H200 (\cref{app:cost}); one run in a fresh process for each baseline, including initialization and compilation, and for \CRAM{} the median of seven complete runs that follow a first run in the same process. 
More details in \cref{app:comparison-protocol}.
}
\label{tab:main}
\fontsize{7.8}{9.4}\selectfont  %
\begingroup
\definecolor{CRAMRow}{HTML}{EEE9F8}
\renewcommand{\TableFamilies}[2]{}
\renewcommand{\TableTime}[2]{\if\relax\detokenize{#2}\relax\makebox[2.6em][r]{#1}\else\makebox[1.9em][r]{#1}\hspace{0.2em}\makebox[0.5em][l]{#2}\fi}
\newcommand{\TableNum}[3]{\if\relax\detokenize{#2}\relax\makebox[2.775em][r]{#1}\else\makebox[1.5em][r]{#1}\makebox[0.75em][l]{.#2}\makebox[0.9em][l]{#3}\fi}
\newcommand{\TableSD}[1]{\makebox[2.9em][l]{\if\relax\detokenize{#1}\relax\else\hspace{0.45em}{\fontsize{6}{7}\selectfont\color{black!50}$\pm$#1}\fi}}
\setlength{\tabcolsep}{4pt}
\renewcommand{\arraystretch}{1.18}
\begin{tabularx}{\linewidth}{Xrrrr@{\hspace{10pt}}rrrr}
\toprule
 & \multicolumn{4}{c}{\TableLabel{Sudoku-Extreme}} & \multicolumn{4}{c}{\TableLabel{Maze-Unique}}\\
\cmidrule(lr){2-5} \cmidrule(lr){6-9}
\textbf{\TableLabel{Model}} & \TableHead{Solved} & \TableHead{Params} & \TableHead{\makebox[4.4em]{FLOPs/puzl}} & \TableHead{Fit time} & \TableHead{Solved} & \TableHead{Params} & \TableHead{\makebox[4.4em]{FLOPs/puzl}} & \TableHead{Fit time}\\
\midrule
\TableHRM{} & 54.62\TableSD{4.18} & \TableNum{27}{0}{M} & \TableSci{2.20}{11} & \TableTime{1.2}{h} & 39.97\TableSD{33.90} & \TableNum{27}{0}{M} & \TableSci{3.01}{12} & \TableTime{7.1}{h}\\
\TableTRM{} & 85.73\TableSD{0.74} & \TableNum{5}{0}{M} & \TableSci{4.11}{11} & \TableTime{2.5}{h} & 92.23\TableSD{13.37} & \TableNum{6}{8}{M} & \TableSci{3.84}{12} & \TableTime{15.6}{h}\\
\TableGRAM{} & 96.10\rlap{$^{\dagger}$}\TableSD{} & \TableNum{9}{1}{M} & \TableSci{8.51}{12} & \TableTime{3.2}{h} & 98.63\TableSD{1.31} & \TableNum{10}{9}{M} & \TableSci{8.00}{13} & \TableTime{23.2}{h}\\
\TableEqR{} & 99.49\TableSD{0.14} & \TableNum{5}{0}{M} & \TableSci{2.11}{14} & \TableTime{2.0}{h} & 93.00\rlap{$^{\dagger}$}\TableSD{} & \TableNum{263}{8}{K} & \TableSci{1.16}{14} & \TableTime{7.2}{h}\\
\TableFRM{} & 99.13\TableSD{0.29} & \TableNum{7}{0}{M} & \TableSci{3.91}{12} & \TableTime{2.9}{h} & 97.27\TableSD{2.80} & \TableNum{8}{0}{M} & \TableSci{4.70}{12} & \TableTime{9.7}{h}\\
\TableTRMPmarked{} & 99.87\TableSD{0.08} & \TableNum{32}{8}{K} & \TableSci{4.57}{12} & \TableTime{9.0}{h} & 99.23\TableSD{0.31} & \TableNum{60}{7}{K} & \TableSci{1.80}{13} & \TableTime{7.6}{h}\\
\TableEBR{} & 99.85\TableSD{0.05} & \TableNum{1}{8}{M} & \TableSci{3.37}{11} & \TableTime{2.8}{h} & 89.80\TableSD{6.35} & \TableNum{1}{8}{M} & \TableSci{1.93}{13} & \TableTime{2.0}{h}\\
\rowcolor{CRAMRow}
\TableCRAM{} & \TableBest{\TableRounded{100.00}}\TableSD{0.00} & \TableBest{\TableNum{7}{7}{K}} & \TableBest{\TableSci{2.02}{9}} & \TableBest{\TableTime{0.04}{s}} & \TableBest{100.00}\TableSD{0.00} & \TableBest{\TableNum{18}{}{}} & \TableBest{\TableSci{8.37}{5}} & \TableBest{\TableTime{0.13}{s}}\\
\TableReference{\TableHand{}} & \TableReference{100.00\TableSD{0.00}} & \TableReference{\TableNum{1}{}{}} & \TableReference{\TableSci{1.08}{8}} & \TableReference{\TableTime{none}{}} & \TableReference{100.00\TableSD{0.00}} & \TableReference{\TableNum{3}{}{}} & \TableReference{\TableSci{8.37}{5}} & \TableReference{\TableTime{none}{}}\\
\bottomrule
\end{tabularx}
\par\smallskip
{\scriptsize\color{black!55}\raggedright $\dagger$ published accuracy, shown where our three-seed reproduction scored lower.\\ $\ddagger$ Hand-set: \CRAM{} with weights set by hand from the task's rules. Params counts unique nonzero values.\\ $\S$ TRMP's abstract also reports a 16k-parameter lattice model that trains in about ten minutes; it pairs the network with a hand-written search (candidate elimination, random commits and restarts). Our reimplementation solves 98.90\% of Sudoku-Extreme after a 13m28s fit (\cref{app:deviations}).\par}
\endgroup
\end{table}
 
\paragraph{Training shapes the energy.}
A common principle of energy-based learning is to lower the energy of observed solutions $\truth{\vect x}\sim P_{\mathrm{data}}$, and raise that of competing answers or \emph{negative samples} $\state{\vect x}^-\sim P_{\vect{\weights}}$ favored by the model, where $P_{\vect{\weights}}(\state{\vect x}^-)\propto\exp[-E_{\vect{\weights}}(\state{\vect x}^-)]$ is the Boltzmann distribution induced by the model's energy \citep{ackley1985,lecun2006energy}.
This objective is depicted in \cref{fig:overview} and mathematically described as

\vspace{0.075cm}
\begin{equation}
\nabla_{\vect{\weights}}\mathcal J_{\mathrm{ML}}(\vect{\weights})
=\term{\mathbb E_{\truth{\vect x}\sim P_{\mathrm{data}}}\!\left[\nabla_{\vect{\weights}}E_{\vect{\weights}}(\truth{\vect x})\right]}{lower energy on data}
-\term{\mathbb E_{\state{\vect x}^-\sim P_{\vect{\weights}}}\!\left[\nabla_{\vect{\weights}}E_{\vect{\weights}}(\state{\vect x}^-)\right]}{raise energy on model samples}.
\label{eq:energy-training}
\end{equation}
\EqNote{Throughout, \truth{green} identifies training labels.}
Importantly, these methods \emph{avoid differentiating through relaxation} (i.e., the looping of \cref{eq:euler}). 
This can significantly reduce fitting cost compared to \BPTT{}, which must propagate gradients through a reasoning trajectory.

Raising energy on model samples (the second term of \cref{eq:energy-training}) can be prohibitively expensive, and several methods approximate it with samples from short Markov chains or relaxations \citep{hinton2002,tieleman2008,movellan1991}.
In this paper, we primarily use a \emph{pseudolikelihood} (PL) objective for its combination of high accuracy, simplicity, and short fitting time on these tasks \citep{besag1975,hyvarinen2006}, though other methods are similarly effective (see \cref{tab:alt-training}).
See \cref{app:pl-ce,app:recipes} for training details. %
\section{Experiments}
\label{sec:results}

In our experiments, a learned energy loses no accuracy against looped models, trains at least 1,000${\times}$ faster, and uses less inference compute (\cref{sec:comparison}).
Its stopping test lets compute follow the difficulty of a puzzle, and its learned weights can be read as the rules of a puzzle (\cref{sec:energy-properties}).
It trains fastest with local objectives, and from few training puzzles (\cref{sec:training-results}).
Finally, we show that simple \CRAM{} energies whose weights are set by hand from the rules of each task solves all tested benchmarks at least as accurately as every trained baseline, using at most three unique weights (\cref{sec:hand-set}).

\paragraph{Tasks.}
\label{sec:task-setup}
We evaluate \CRAM{} on four reasoning tasks:
Sudoku-Extreme \citep{hrm2025} (a set of hard $9\times9$ Sudoku puzzles),
Maze-Unique \citep{eqr2026} (a set of $30\times30$ mazes whose answer is the unique path between two endpoints), N-Queens, and Graph Coloring \citep{gram2026,ebr2026} (tasks that have many valid answers per input).
Every task trains the pairwise energy of \cref{eq:core}, and we specify how each task informs the structure of \CRAM{}'s weights in \cref{tab:choices}.
In no task is the energy told the rules of a puzzle.

\paragraph{Baselines.}
We test against HRM \citep{hrm2025}, Tiny Recursive Model (TRM) \citep{trm2025}, Generative Recursive Reasoning (GRAM) \citep{gram2026}, Equilibrium Reasoners (EqR) \citep{eqr2026}, Flow Reasoning Models (FRM) \citep{frm2026}, Tiny Recursive Message Passing (TRMP) \citep{trmp2026} and Energy Based Reasoners (EBR) \citep{ebr2026}.
Where available, we retrain each baseline with its authors' code and recipe on one H200 for timing and score each baseline with its published inference setting.\footnote{The training code of TRMP is not released, so we reimplement TRMP from the recipe in its paper (\cref{app:deviations}).}
We score every method on the same test set of each task, and no method chooses a setting or a checkpoint on the test set (\cref{app:comparison-protocol}).
\CRAM{} and each baseline trains with three seeds, and we report the mean over the three runs.
We compare only against models that produce the answer by their own recurrent computation. 
Methods that pass learned costs to an external exact solver \citep{brouard2020,defresne2023,epll2025} are outside the scope of this work.

\paragraph{Training \CRAM{}.}
\CRAM{} weights are initialized to zero and are trained to minimizes the PL objective of \cref{eq:pl} at the labeled solutions using with SGD with momentum and a cosine schedule.
The priors of \cref{tab:choices} enter as shared weights or, on Sudoku-Extreme, as digit relabeling of the training boards.
We take the final trained model and do not select a best performing checkpoint when reporting \CRAM{}'s results (\cref{app:recipes}).
Training runs finish by computing the stopping threshold from the training solutions (their highest energy plus a small tolerance); reported fit times include this step.

\begin{figure}[t]
\centering
\includegraphics[width=\linewidth]{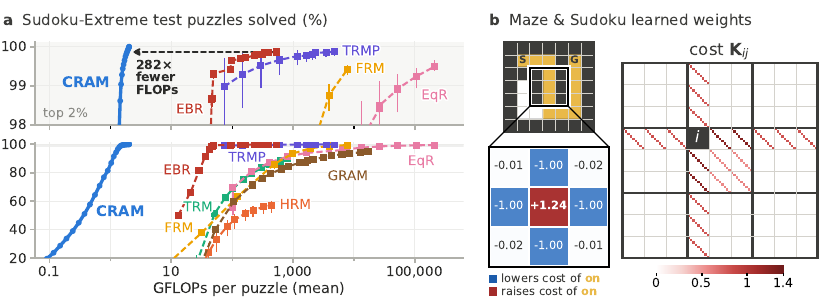}
\caption{\textbf{(a)~\CRAM{} matches each looped model's highest accuracy using at least 282${\times}$ fewer inference FLOPs.} 
Each point is one method showing what fraction of a 10,000 Sudoku-Extreme test puzzle was solved within a given inference budget, reported as the mean accuracy over three training runs with error bars spanning the lowest to the highest run.
The arrow joins the most accurate point of EBR, the cheapest looped model, to the cheapest \CRAM{} point that solves at least as many puzzles. 
The upper panel zooms into the top 2\% shaded are of the lower panel.
\CRAM{}, TRMP and EBR points include early stopping, and FLOPs include the full reasoning search including every search chain (see \cref{app:opcounts} for details).
\textbf{(b)~\CRAM{} learns weights that describe the reasoning task.} Maze-Unique learns 3${\times}$3 shared weights, shown in units of the mean coupling to the four side neighbours, that describe a dead-end pruning algorithm for solving Maze-Unique, by exclusively observing valid maze solutions. 
On Sudoku-Extreme, \CRAM{} learns correct sudoku costs despite receiving no positional constraint graph. Trained weights for sudoku and maze correspond to the \CRAM{} model reported in \cref{tab:main}.}

\label{fig:compute}
\label{fig:sudoku-couplings}
\end{figure}
 
\subsection{Comparison with looped models}
\label{sec:comparison}

\paragraph{High accuracy at small sizes and fast training.}
Does a \CRAM{} energy that learns from solved puzzles reach the accuracy of looped models, and at what cost?
\Cref{tab:main} compares \CRAM{} with the seven looped baselines on Sudoku-Extreme and Maze-Unique.
On both tasks, \CRAM{} is the most accurate method, solving 99.996\% of the Sudoku-Extreme test puzzles and every Maze-Unique test maze.
The most accurate looped model, TRMP, solves 99.87\% and 99.23\%.
\CRAM{} reaches this accuracy with thousands of times less training time than every looped model.
All results are trained from examples except the gray rows of \cref{tab:main}, which are a reference for an optimal \CRAM{} energy that is coded by hand from the puzzles explicit constraints (\cref{sec:hand-set}).

\begin{table}[!t]
\centering
\caption{\textbf{\CRAM{} matches larger baselines on N-Queens and graph coloring.}
Both N-Queens on $8{\times}8$ and $10{\times}10$ boards and graph coloring of graphs with 8 and 10 vertices are evaluated on GRAM's test sets.
All numbers report mean {\color{black!50}$\pm$stdev} over three training seeds; the best learned value of each row is in bold.
\TableCaptionLabel{Acc.}: test inputs whose kept answer is correct (\%); $N{=}1$ keeps one sample, and $N{=}20$ keeps the majority vote of 20 samples for GRAM and the lowest-energy sample of 20 for EBR and \CRAM{}; TRM and HRM are deterministic, so their two values are equal. \TableCaptionLabel{Cov@20}: share of each input's solutions found by 20 samples (\%).
\TableCaptionLabel{Conflict}: conflicting edges of the kept answer, summed over the test graphs.
\TableCaptionLabel{Params}: as in \cref{tab:main}; each task trains its own model.
\TableCaptionLabel{Fit time}: training seed 0 on one H200; one run for each baseline, and for \CRAM{} the median of seven runs that follow a first run in the same process (\cref{app:nqueens}).}
\label{tab:nq-gc}
\fontsize{7.5}{8.8}\selectfont  %
\begingroup
\definecolor{CRAMRow}{HTML}{EEE9F8}
\renewcommand{\TableFamilies}[2]{}
\newcommand{\MSsd}[1]{\makebox[2.1em][l]{\hspace{0.15em}{\fontsize{5.5}{6}\selectfont\color{black!50}$\pm$#1}}}
\setlength{\tabcolsep}{3pt}
\renewcommand{\arraystretch}{1.0}
\begin{tabular*}{\linewidth}{@{\extracolsep{\fill}}c l rrrr>{\columncolor{CRAMRow}}r r@{}}
\toprule
\textbf{\TableLabel{Task}} & & \multicolumn{1}{r}{\TableHRM{}} & \multicolumn{1}{r}{\TableTRM{}} & \multicolumn{1}{r}{\TableGRAM{}} & \multicolumn{1}{r}{\TableEBR{}} & \multicolumn{1}{>{\columncolor{CRAMRow}}r}{\TableCRAM{}} & \multicolumn{1}{r}{\TableReference{\TableHand{}}}\\
\midrule
\multirow{5}{*}{\TableLabel{\shortstack{N-Queens\\$8{\times}8$}}} & \TableLabel{Acc., $N{=}1$} & 84.41\MSsd{2.2} & 67.39\MSsd{6.8} & 99.44\MSsd{0.2} & \TableBest{100.00}\MSsd{0.0} & 99.95\MSsd{0.0} & \TableReference{100.00\MSsd{0.0}}\\
 & \TableLabel{Acc., $N{=}20$} & 84.41\MSsd{2.2} & 67.39\MSsd{6.8} & \TableBest{100.00}\MSsd{0.0} & 99.95\MSsd{0.1} & \TableBest{100.00}\MSsd{0.0} & \TableReference{100.00\MSsd{0.0}}\\
 & \TableLabel{Cov@20} & 58.8\MSsd{0.6} & 51.4\MSsd{3.1} & 97.9\MSsd{0.2} & \TableBest{98.1}\MSsd{0.2} & 97.9\MSsd{0.1} & \TableReference{98.4\MSsd{0.0}}\\
 & \TableLabel{Params} & 27.3M\makebox[2.1em]{} & 6.8M\makebox[2.1em]{} & 10.9M\makebox[2.1em]{} & 7.1M\makebox[2.1em]{} & \TableBest{113}\makebox[2.1em]{} & \TableReference{2\makebox[2.1em]{}}\\
 & \TableLabel{Fit time} & 59\,min\makebox[2.1em]{} & 1.3\,h\makebox[2.1em]{} & 1.3\,h\makebox[2.1em]{} & 3.2\,h\makebox[2.1em]{} & \TableBest{2.9\,s}\makebox[2.1em]{} & \TableReference{none\makebox[2.1em]{}}\\
\midrule
\multirow{5}{*}{\TableLabel{\shortstack{N-Queens\\$10{\times}10$}}} & \TableLabel{Acc., $N{=}1$} & 37.80\rlap{$^{\dagger}$}\makebox[2.1em]{} & 17.50\rlap{$^{\dagger}$}\makebox[2.1em]{} & 88.91\MSsd{3.1} & 99.31\MSsd{0.3} & \TableBest{99.45}\MSsd{0.1} & \TableReference{99.99\MSsd{0.0}}\\
 & \TableLabel{Acc., $N{=}20$} & 37.80\rlap{$^{\dagger}$}\makebox[2.1em]{} & 17.50\rlap{$^{\dagger}$}\makebox[2.1em]{} & 99.98\MSsd{0.0} & \TableBest{99.99}\MSsd{0.0} & \TableBest{99.99}\MSsd{0.0} & \TableReference{100.00\MSsd{0.0}}\\
 & \TableLabel{Cov@20} & 10.7\MSsd{4.0} & 3.3\MSsd{3.6} & 92.9\MSsd{0.3} & 92.5\MSsd{0.7} & \TableBest{93.6}\MSsd{0.1} & \TableReference{95.5\MSsd{0.0}}\\
 & \TableLabel{Params} & 27.3M\makebox[2.1em]{} & 6.8M\makebox[2.1em]{} & 10.9M\makebox[2.1em]{} & 7.1M\makebox[2.1em]{} & \TableBest{181}\makebox[2.1em]{} & \TableReference{2\makebox[2.1em]{}}\\
 & \TableLabel{Fit time} & 4.0\,h\makebox[2.1em]{} & 5.0\,h\makebox[2.1em]{} & 5.2\,h\makebox[2.1em]{} & 7.6\,h\makebox[2.1em]{} & \TableBest{6.8\,s}\makebox[2.1em]{} & \TableReference{none\makebox[2.1em]{}}\\
\midrule
\multirow{5}{*}{\TableLabel{\shortstack{Graph coloring\\8 vertices}}} & \TableLabel{Conflict, $N{=}1$} & 139.7\MSsd{10.7} & 118.0\MSsd{15.5} & 3.3\MSsd{2.5} & \TableBest{0.0}\MSsd{0.0} & \TableBest{0.0}\MSsd{0.0} & \TableReference{0.0\MSsd{0.0}}\\
 & \TableLabel{Conflict, $N{=}20$} & 139.7\MSsd{10.7} & 118.0\MSsd{15.5} & 3.3\MSsd{1.2} & 2.7\MSsd{4.6} & \TableBest{0.0}\MSsd{0.0} & \TableReference{0.0\MSsd{0.0}}\\
 & \TableLabel{Cov@20} & 20.6\MSsd{0.4} & 21.8\MSsd{0.8} & 87.9\MSsd{0.6} & \TableBest{88.9}\MSsd{0.1} & \TableBest{88.9}\MSsd{0.5} & \TableReference{90.6\MSsd{0.6}}\\
 & \TableLabel{Params} & 27.3M\makebox[2.1em]{} & 6.8M\makebox[2.1em]{} & 10.9M\makebox[2.1em]{} & 7.1M\makebox[2.1em]{} & \TableBest{869}\makebox[2.1em]{} & \TableReference{1\makebox[2.1em]{}}\\
 & \TableLabel{Fit time} & 41\,min\makebox[2.1em]{} & 54\,min\makebox[2.1em]{} & 53\,min\makebox[2.1em]{} & 4.5\,h\makebox[2.1em]{} & \TableBest{1.3\,s}\makebox[2.1em]{} & \TableReference{none\makebox[2.1em]{}}\\
\midrule
\multirow{5}{*}{\TableLabel{\shortstack{Graph coloring\\10 vertices}}} & \TableLabel{Conflict, $N{=}1$} & 182.7\MSsd{17.8} & 169.0\MSsd{20.3} & 52.7\MSsd{82.6} & \TableBest{0.3}\MSsd{0.6} & \TableBest{0.3}\MSsd{0.6} & \TableReference{0.0\MSsd{0.0}}\\
 & \TableLabel{Conflict, $N{=}20$} & 182.7\MSsd{17.8} & 169.0\MSsd{20.3} & 52.7\MSsd{83.4} & 0.7\MSsd{1.2} & \TableBest{0.0}\MSsd{0.0} & \TableReference{0.0\MSsd{0.0}}\\
 & \TableLabel{Cov@20} & 6.2\MSsd{0.9} & 6.9\MSsd{0.5} & 37.8\MSsd{24.7} & \TableBest{54.8}\MSsd{0.2} & 52.6\MSsd{0.6} & \TableReference{55.2\MSsd{0.4}}\\
 & \TableLabel{Params} & 27.3M\makebox[2.1em]{} & 6.8M\makebox[2.1em]{} & 10.9M\makebox[2.1em]{} & 7.1M\makebox[2.1em]{} & \TableBest{475}\makebox[2.1em]{} & \TableReference{1\makebox[2.1em]{}}\\
 & \TableLabel{Fit time} & 1.7\,h\makebox[2.1em]{} & 2.1\,h\makebox[2.1em]{} & 2.2\,h\makebox[2.1em]{} & 7.6\,h\makebox[2.1em]{} & \TableBest{0.3\,s}\makebox[2.1em]{} & \TableReference{none\makebox[2.1em]{}}\\
\bottomrule
\end{tabular*}
\endgroup
 \par\vspace{3pt}
{\fontsize{6.8}{8.2}\selectfont\color{black!55}\raggedright N-Queens Cov@20 follows EBR's evaluator, normalizing over every valid solution per puzzle. GRAM reports a different normalizer.\par
$\dagger$ published accuracy, shown where our three-seed reproduction scored lower.\par
$\ddagger$ Hand-set: \CRAM{} with weights set by hand from the task's rules. Params counts unique nonzero values.\par}
\end{table}

\paragraph{Efficient inference.}
Each method has a setting that allows us to vary its maximum allowed inference budget: the number of reasoning steps (HRM, TRM, FRM), restarts (EqR), sampled trajectories (GRAM, EBR), rollouts (TRMP), or for \CRAM{}, the number of allowed search moves.
By sweeping across allowed inference budgets of each method on 10,000 Sudoku-Extreme test puzzles, we can get a simple ``scaling law'' plot for inference compute, as shown in \Cref{fig:compute}a.
On this plot, \CRAM{} uses between 282${\times}$ (against EBR) and 128,000${\times}$ (against EqR) fewer FLOPs than the looped models when compared at each baseline's most accurate setting.
\CRAM{} reaches EBR's top accuracy (99.86\%) using $1.9\times10^9$ FLOPs per puzzle vs EBR's $5.4\times 10^{11}$ FLOPs.
At max accuracy, \CRAM{} solves 99.997\% of the puzzles with a mean of $2.0\times10^9$ FLOPs per puzzle.
Note that \CRAM{} never returns a wrong answer, it only ``misses'' where a search reaches its budget before any board passes the stopping threshold.

\paragraph{Diverse valid solutions.}
N-Queens and graph coloring test whether the same energy finds many different answers. 
Since both have many solutions per input, we follow GRAM and EBR \citep{gram2026,ebr2026} and report two scores in \cref{tab:nq-gc}: the accuracy of a single answer (out of $N=1$ or $N=20$ chains), and a valid-diversity score over 20 samples (Cov@20). 
\CRAM{}'s learned energy performs competitively to the best looped baselines across all metrics, despite training in seconds. 
Note that every Cov@20 value here comes from EBR's evaluator, since GRAM publishes values from a different normalizer.

\subsection{Advantages of explicit energy}
\label{sec:energy-properties}

\begin{figure}[t]
\centering
\includegraphics[width=\linewidth]{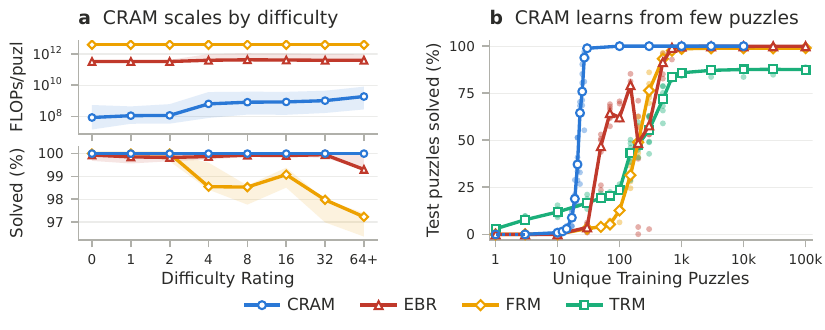}
\caption{\textbf{(a)~\CRAM{}'s compute grows gently with puzzle difficulty.}
The 10,000 Sudoku-Extreme test puzzles of \cref{fig:compute} are binned by difficulty rating provided by the dataset, where each tick gives the lowest rating of its bin (0, 1, 2--3, 4--7, 8--15, 16--31, 32--63, and 64+).
FRM uses static compute for every puzzle, whereas both \CRAM{} and EBR implement early stopping.
EBR's early stop barely lowers its compute on easy puzzles: its median stays between 3.3 and $4.2\times10^{11}$ FLOPs in every bin, near its minimum cost.
Top: median FLOPs per puzzle, with bands spanning the 10th to 90th percentiles.
Bottom: test accuracy, median over three training runs, with bands spanning the runs.
\textbf{(b)~\CRAM{} is data efficient, learning Sudoku using an order of magnitude fewer examples.}
Shown is model accuracy on the same 10,000 test puzzles vs. the number of unique training puzzles seen in the data (each data point is augmented with the same 99 augmentations).
Points connect the median value across 6 training runs.
}
\label{fig:difficulty-data}
\end{figure}

\paragraph{Compute that adapts to difficulty.}
\CRAM{}'s search stops at the first board whose energy is at most the threshold fitted from the training solutions, which lets \CRAM{} dynamically adjust its compute to follow problem difficulty.
When we bin the 10,000 test puzzles of \cref{fig:compute}a by the difficulty rating of the dataset (\cref{fig:difficulty-data}a), the median compute of \CRAM{} grows 22${\times}$ from the easiest bin to the hardest.
In contrast, FRM spends the same compute for every puzzle and EBR, whose gradient-norm stopping rule enables early stopping for Sudoku, barely changes compute needs across puzzle difficulty.
Note that the stopping test of \CRAM{} is also reliable: it never stops at an incorrect board, whereas the halting rules of HRM, TRM, EqR, TRMP and EBR each stop at an incorrect board on 0.05--3.54\% of the puzzles that they stop on (\cref{tab:halting}).

\paragraph{Inspectable learned weights.}
\label{sec:what-learned}
The nature of these logic puzzles mean we can read the rules of the puzzle directly from \CRAM{}'s learned weights (\cref{fig:sudoku-couplings}b). 
Despite receiving no task-informed constraints for Sudoku, in every training run \CRAM{} learned to put positive equal-digit costs on exactly the 810 cell pairs that share a row, column or box, leaving all other costs zero.
On Maze-Unique, the 18 learned costs make each move of the search turn off one dead-end cell, which is dead-end filling \citep{pullen_maze_algorithms}, a local rule that solves every perfect maze (\cref{app:maze-unique}).

\subsection{Training the energy}
\label{sec:training-results}

\begin{figure}[!t]  %
\centering
\includegraphics[width=\linewidth]{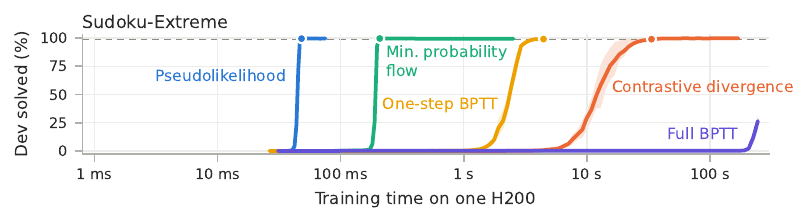}
\caption{\textbf{Local methods that avoid sampling the energy train significantly faster.} We train \CRAM{} with five objectives, the four named below and \textcolor{AltBPTTone}{one-step \BPTT{}} \citep{hrm2025}, and plot the mean solve rate on 512 Sudoku-Extreme development puzzles over 20 training runs against training time.
Bands span the 25th to 75th percentile of the runs.
The mean reaches 99\% (circles) after 48\,ms with \textcolor{AltPL}{pseudolikelihood} \citep{besag1975} and 0.21\,s with \textcolor{AltMPF}{minimum probability flow} \citep{sohldickstein2011}, against 4.4\,s with one-step \BPTT{} and 33\,s with \textcolor{AltCD}{contrastive divergence} \citep{hinton2002}.
\textcolor{AltBPTT}{Full \BPTT{}} \citep{werbos1990} ends at a mean of 27\% after 245\,s, still rising at the end of its schedule; with 3,840,000 updates (about 5\,h) it solves 48.8\% of test puzzles.
Training times include fitting the stopping threshold from the training solutions (24--75\,ms) and exclude each run's one-time start-up (0.08--0.30\,s), as do the T$_{99}$ times of \cref{tab:alt-training}, which compares twelve methods on Sudoku-Extreme and Maze-Unique; \cref{app:alt-training} gives details of each objective.}
\label{fig:alt-training}
\end{figure}

\paragraph{Characterizing fast training objectives.}
All results above use pseudolikelihood (PL), but the energy of \cref{eq:core} accepts many other training methods.
We compare three classes of training methods on Sudoku-Extreme to see which methods train \CRAM{} the fastest (wall-clock time on a single H200).
\emph{Local objectives} such as PL are evaluated directly at the labeled solution, with no unrolled trajectory during training.
\emph{Sampling-based methods} draw the negative samples of \cref{eq:energy-training} by unrolling the model, but no differentiation is taken through those unrolled trajectories.
A third class \emph{differentiates through the relaxation} of \cref{eq:flow}. This is how looped models are traditionally trained, using truncated or full BPTT, but it also includes methods that approximate BPTT such as implicit differentiation \citep{pineda1987,deq2019}.
\cref{fig:alt-training} shows that local objectives like pseudolikelihood and min probability flow~\cite{sohldickstein2011} train fastest, followed by short, truncated \BPTT{}. Contrastive divergence follows, and full \BPTT{} (training through all 10 unrolled relaxation steps) fails to solve more than 48.8\% after 5\,h of training.
Full comparison of results are in \cref{tab:alt-training} of  \cref{app:alt-training}.

\paragraph{Data efficient learning.}
\label{sec:data-efficiency}
\CRAM{} is data efficient. \cref{fig:difficulty-data}b explores how many unique Sudoku-Extreme training puzzles are needed to solve a 10k Sudoku-Extreme test set.
Each method is provided the same unique puzzles, symmetrically augmented an additional 99 times.
On this dataset, \CRAM{} needs an order of magnitude fewer puzzles to reach 98\% accuracy than looped models, using only 30 unique puzzles to EBR's 700. Interestingly, TRM solves the most test puzzles in the extremely data-limited regime ($<10$ puzzles), but it scales much slower than both energy-based methods. 
In the low-data regime, \CRAM{} converged to many incorrect solutions.

\paragraph{Hand-set energies perform best.}
\label{sec:hand-set}
In \cref{tab:main,tab:nq-gc}, we compare a \CRAM{} energy whose weights were set by hand using the rules of a task (\cref{app:hand-set}).
Under the same search as learned \CRAM{}, the hand-set energies solve every Sudoku, Maze, N-Queens, and Graph Coloring test puzzle, using up to 19${\times}$ fewer inference FLOPs than a learned energy on Sudoku.
On both N-Queens and Graph Coloring, hand-set \CRAM{} is the top performing model across all metrics and board sizes. 
A high score on these benchmarks therefore only shows that a model has learned a few pairwise rules between answer cells---a gentle caution against throwing larger looping models at these reasoning problems.

\section{Discussion}
\label{sec:limitations}

Recent work has suggested the looping provides an alternative to chain-of-thought on reasoning tasks. In this work, we have used the formal framework of energies to provide a concrete basis for looped models, in which convergence is guaranteed, computations can adapt to problem difficulty, and BPTT can be avoided. We introduced a specific trainable energy, \CRAM{}, and have shown that it is able to learn a suite of commonly used reasoning benchmarks three orders of magnitude faster than other looped methods. 

We note that all results with fit times are wall-clock times on a shared cluster, so they vary with concurrent work on a node.
For any model that trains on the order of seconds or minutes, where compilation could meaningfully affect reported training times, we report warm fit times that avoid one-time startup costs, such as kernel loading and compilation (see \cref{app:cost}). Longer runs report cold-fit times. 

The puzzles evaluated in this work---Sudoku-Extreme \citep{hrm2025}, Maze-Unique \citep{eqr2026}, and the N-Queens and graph-coloring sets of \citet{gram2026}---belong to a category of logic puzzles that \CRAM{}'s pairwise energy can represent exactly \citep{hopfieldtank1985,hopfieldsudoku2008}. 
In fact, \Cref{tab:main} shows that a hand-set \CRAM{} energy with one unique weight value solves every Sudoku-Extreme test puzzle, and one with three values solves every Maze-Unique test maze.
High accuracy on these logic benchmarks therefore does not indicate that a model can scale to harder tasks such as ARC-AGI-1 \citep{hrm2025,chollet2019},
where the answer depends on quantities that the model must compute from the visible board, such as each cell's distance to the goal in Maze-Hard.
In \cref{sec:maze-hard-latents}, we show how a simple latent extension of \CRAM{} can learn nearly all of the 1,000 Maze-Hard test mazes, but learning such latent quantities in general and without foreknowledge of the task's design remains an important open research direction to scale these energies to harder reasoning tasks.

Building efficient reasoning models that are applicable to general tasks is a central challenge in AI research.  
Our results indicate that energy is a powerful framework for reasoning on constraint puzzles, offering almost 300${\times}$ faster inference and 1,000${\times}$ faster training than looped models with no loss in accuracy.
Though the energy studied here is not a final answer for all reasoning tasks, it showcases the incredible benefit that an explicit energy provides, and provides a promising path towards data-efficient general reasoning capabilities.
 \label{main-end}
\ifdefined\SubmissionCopy\else
\section*{Acknowledgments}
We thank Ales Kubicek, Kai Sandbrink, Andrea Caciolai, Fraser Greenlee, Liangzu Peng, Zihan Yan, Sukrit Sumant, Alec Helbling, Adrian Rodriguez, Ruairidh Battleday, and Zeb Kurth-Nelson for insightful discussions and feedback.
\fi
\clearpage
\bibliographystyle{plainnat}
\bibliography{references}
\clearpage
\appendix
\section{Stopping rules of looped models and \CRAM{}}
\label[appendix]{app:halting}
HRM, TRM, EqR and TRMP learn early stopping with a separate network \citep{hrm2025,trm2025,eqr2026,act2016}.
EBR stops a sample when the norm of its energy gradient falls below 0.2 \citep{ebr2026}.
\CRAM{} is different, stopping at the first board whose energy is at most a threshold fitted from the training solutions.
\Cref{tab:halting} shows that only \CRAM{} never stops at an incorrect board --- all other early stopping methods do. 
Note that EBR's stop does not work on Maze-Unique since its gradient norm stays near 2 even at correct mazes. EBR's published maze protocol runs a fixed number of segments instead.

\begin{table}[ht]
\centering
\caption{\textbf{\CRAM{} never halts at an incorrect board.} Each looped model runs one rollout in its halting mode, capped at its trained depth; EBR's first sample stops at a gradient norm below 0.2 or after 64 segments, as in its code. Halted: puzzles on which the rule stopped before the cap. Wrong halts: halted puzzles whose board is incorrect (\% of halted). Full test sets (412,786 puzzles and 1,000 mazes), pooled over three training runs.}
\label{tab:halting}
\footnotesize
\definecolor{CRAMRow}{HTML}{EEE9F8}  %
\setlength{\tabcolsep}{5pt}  %
\begin{tabular}{@{}lrrrrrr@{}}
\toprule
 & \multicolumn{3}{c}{Sudoku-Extreme} & \multicolumn{3}{c}{Maze-Unique}\\
\cmidrule(lr){2-4} \cmidrule(lr){5-7}
Stopping rule & Halted (\%) & Wrong halts (\%) & Solved (\%) & Halted (\%) & Wrong halts (\%) & Solved (\%)\\
\midrule
HRM (halting head) & 54.5 & 0.62 & 54.7 & 21.4 & 0.16 & 40.3\\
TRM (halting head) & 85.8 & 2.37 & 83.8 & 99.9 & 3.54 & 96.4\\
EqR (halting head) & 87.5 & 1.61 & 86.1 & 66.0 & 1.06 & 65.3\\
TRMP (halting head) & 93.8 & 0.05 & 93.8 & 91.2 & 0.73 & 99.2\\
EBR (gradient-norm stop) & 99.35 & 0.21 & 99.31 & \multicolumn{3}{c}{\textcolor{black!45}{no stopping rule}}\\
\rowcolor{CRAMRow}
\CRAM{} (energy threshold) & 99.996 & 0.00 & 99.996 & 100.00 & 0.00 & 100.00\\
\bottomrule
\end{tabular}
\end{table}
 
\FloatBarrier
\section{Compute}
\label[appendix]{app:compute}
Every experiment in this paper runs on one NVIDIA H200 GPU.
Training jobs used GPUs on unreserved nodes of a shared cluster, so other jobs could use a node's other GPUs and CPU cores which could affect reproducibility of timing a fit.
Wherever CPU times are reported, they use  one thread of an Intel Xeon Platinum 8468 CPU.
\CRAM{}'s search moves run in Triton kernels with one GPU warp per chain.
A move changes one cell, so the kernel keeps the field of every cell in memory and updates only the fields of the cells with a nonzero coupling to the moved cell.
The code, data preparation, run records and exported results behind every table and figure are at \url{https://github.com/Prefrontal-AI/CRAM}.

\FloatBarrier
\section{Experimental protocols}
\label[appendix]{app:protocols}
\label[appendix]{app:recipes}

\subsection{Shared protocol}
\label[appendix]{app:comparison-protocol}
Every method, \CRAM{} and each baseline, is trained three times under different seeds, and each trained model is scored once on the test set.
Where a published accuracy was selected on the test set, we say so.
The test sets have 412,786 puzzles (Sudoku-Extreme), 1,000 mazes (Maze-Unique), 1,240 and 18,466 rows (N-Queens, $8\times8$ and $10\times10$) and 255 and 192 graphs (graph coloring, 8 and 10 vertices).
When reproducing baselines, whenever our three-seed mean accuracy is more 2 points below their published accuracy, a table shows the published value (dagger), but we still count that baseline's parameters and FLOPs ourselves.
A baseline's Params is the parameter count of its network: we do not check its weights for zeros, so it is an upper bound on the nonzero parameters that \CRAM{}'s count reports.

\CRAM{} inference is a focused Gibbs search with four chains (Sudoku-Extreme) or with $N$ chains ($N{=}1$ or 20 in \cref{tab:nq-gc}; N-Queens and graph coloring), and zero-temperature single-cell moves (Maze-Unique).
In general, we use each baseline's published inference setting when solving a task. 

\begin{itemize}
\item \textbf{TRMP}: the most confident of 64 noisy rollouts (the ACT model in \cref{tab:main}).
\item \textbf{TRMP's lattice model} (the \S{} footnote of \cref{tab:main}): 64 search chains, each restarted on a detected conflict, for at most 1,000 steps.
\item \textbf{EqR}: depth 64 with 128 noisy restarts, answer chosen by the model's convergence score \citep{eqr2026}.
\item \textbf{GRAM}: majority vote of 20 sampled trajectories of 16 steps.
\item \textbf{EBR}: the lowest-energy of five samples; OR when its energy-gradient norm falls below 0.2, or after 64 segments (Sudoku Extreme); OR after 16 updates (Maze-Unique)
\end{itemize}

\subsection{Operation counts and fit time}
\label[appendix]{app:cost}
\label[appendix]{app:opcounts}

\textbf{Reported FLOPs} count the operations along all their sampling trajectories, stopping tests, selection amongst samples, and decoding; it excludes losses and metrics computed from labels.

Every FLOP count in this paper, for \CRAM{} and every baseline, uses the following convention:

\begin{itemize}
\item A matrix product counts 2 FLOPs per multiply-add.
\item Every elementwise addition, subtraction, multiplication (also by a 0/1 mask), division, negation and absolute value counts 1, on floats or integers.
\item Every exp, log, square root, sigmoid, tanh and erf counts 1.
\item Comparisons, selections, min, max, argmax, mask logic, indexing, data movement, casts and random-number generation count 0. Arithmetic applied to a random number (scaling noise, a Gumbel transform) counts.
\end{itemize}

For the neural baselines, a PyTorch counter at the operator-dispatch level applies these rules to every operator during inference, including gradient computations.
\CRAM{}'s moves use Triton kernels that are invisible to the counter, so count their transcription line by line. \CRAM{}'s code outside the kernels is counted directly.
A \CRAM{} search counts FLOPs up until a move first passes its stopping test.

\textbf{Reported fit times} are the wall clock of one complete fit.
The clock starts when the training data are in host memory and stops when the inference-ready model is read, including its stopping statistics.
In total, it includes: transfer to the device, initialization, augmentation, every forward, backward and optimizer step, compilation and CUDA-graph capture with their warm-up, and fitting stopping thresholds and focus statistics from the training solutions.
It excludes reading files, evaluation, saving and module imports.

We report the warm time for \CRAM{}, because its one-time start-up (0.2--0.9\,s on one H200) is a large part of a fit that takes milliseconds, but a negligible part of a baseline fit that takes hours. 

\begin{itemize}
\item \textbf{Cold}: the first fit in a fresh process, including one-time kernel loading, compilation and graph capture. CUDA context creation (0.4--0.7\,s) is measured separately and excluded.
\item \textbf{Warm}: the median of seven further fits in the same process.
\end{itemize}

It was not within our compute budget to compute warm-starts for long baseline runs.
Thus, each baseline time is one cold fit on one H200 with an empty compilation cache.
Its start-up (compilation, kernel loading and the first update) takes 6\,s to 16\,min, at most 2.2\% of the fit (3.5\% for TRMP on Maze-Unique).

\subsection{\CRAM{} recipes}
\label[appendix]{app:cram-recipes}

\Cref{tab:recipes} lists the training and search settings of every \CRAM{} fit.
Two fits add a step that the table does not list.
The Sudoku-Extreme fit applies a fresh random digit relabeling to the training boards at each update.
The N-Queens fit divides the gradient of each offset cost by a per-offset preconditioner (\cref{app:nqueens}).

\begin{table}[h]
\centering
\caption{\textbf{Every \CRAM{} fit is a few dozen to a few thousand SGD updates of pseudolikelihood, and every search is a single-cell local search.} All fits start at zero and use SGD with momentum 0.5 and a cosine schedule; the L1 term is a soft-threshold after each step. Values in parentheses are for the larger N-Queens board or graph. $\bar h$: mean positive learned cost; $u$: the N-Queens temperature unit (\cref{app:nqueens}); $\tau$: highest energy of a training solution.}
\label{tab:recipes}
\footnotesize
\setlength{\tabcolsep}{3pt}
\renewcommand{\arraystretch}{1.1}
\begin{tabularx}{\linewidth}{@{}>{\bfseries}l >{\raggedright\arraybackslash}X >{\raggedright\arraybackslash}X >{\raggedright\arraybackslash}X >{\raggedright\arraybackslash}X@{}}
\toprule
 & Sudoku-Extreme & Maze-Unique & N-Queens $8\times8$ ($10\times10$) & Graph coloring, 8 (10) vertices\\
\midrule
\rowcolor{black!8}\multicolumn{5}{@{}l}{\normalfont\strut\textit{Training}}\\
Training rows & 1,000 & 1,000 & 7,224 (108,234) & 7,002 (13,465)\\
Loss & PL + L1, $\lambda=3.33\times10^{-4}$ & PL & PL & PL + L1, $\lambda=10^{-4}$ ($10^{-3}$)\\
Updates & 32, full batch & 20, batches of 64 & 4,000, full batch & 1,024 (256), full batch\\
Learning rate & 540 & 1 & 30 & 10\\
\addlinespace[3pt]
\rowcolor{black!8}\multicolumn{5}{@{}l}{\normalfont\strut\textit{Search}}\\
Start & uniform random & every open cell on & given queens only & uniform random\\
Moves & focused Gibbs & $T=0$ single-cell & focused Gibbs & focused Gibbs\\
Temperature & $0.40\bar h\to0.31\bar h$ & 0 & $u$ & $2\bar h$ ($3\bar h$)\\
Chains & 4 & 1 & 1 or 20 & 1 or 20\\
Stopping test & energy $\le\tau+10^{-3}$ & fixed point & local margin & every term below its training maximum\\
Budget & $6.3\times10^{11}$ FLOPs & 20,000 moves & 20,000 moves & 10,000 moves\\
\bottomrule
\end{tabularx}
\end{table}

\subsubsection{Sudoku-Extreme}
\label[appendix]{app:sudoku}
Sudoku-Extreme \citep{hrm2025} (\texttt{sapientinc/sudoku-extreme}) has an official training split and a test split of 422,786 puzzles.
\CRAM{} trains on the three 1,000-puzzle training splits of FRM \citep{frm2026}, one per seed, which FRM draws from the training split balanced over its difficulty-rating bins.
Our development set is 10,000 puzzles of the test split, and our test set is the other 412,786.
The cells are opaque indices: \CRAM{} is told neither the rows, columns and boxes nor any structure of the digits.
Each pair of cells has a free $9\times9$ cost block $\vect{\paircost}_{ij}=\vect{\paircost}_{ji}^{\top}$ and each cell a free input cost $\vect{\inputcost}_i\in\R^9$ (263,169 parameters), clue cells are clamped, and training follows \cref{tab:recipes}.

FRM, EqR and TRMP train on the same splits.
HRM, TRM, GRAM and EBR keep the data pipeline of their own code: HRM, TRM and GRAM train on one 1,000-puzzle draw from TRM's data builder (the command of the authors' README, which fixes no random seed), with 1,000 stored symmetry variants per puzzle, and EBR trains on its own draw (seed 42).
\Cref{fig:difficulty-data}b trains \CRAM{}, TRM, EBR and FRM on identical rows.

For \CRAM{} each sudoku puzzle runs four chains from random boards.
A move picks a uniformly random free cell whose local energy exceeds its training-solution mean by more than $0.45\bar h$ (any free cell if there is none) and samples its digit from its belief.
The temperature falls geometrically from $0.40\bar h$ to $0.31\bar h$ over 524,288 moves and then restarts, and the cached fields are recomputed exactly every 4,096 moves.
A puzzle stops at the first board of any chain with energy at most $\tau+10^{-3}$, where $\tau$ is the highest energy of a training solution of its split ($-1.7\times10^{-4}$ to $-3\times10^{-6}$), and otherwise returns its lowest-energy board at the budget.
We tuned the temperatures and the stopping threshold on development puzzles.
On the test set, no search stopped at an incorrect board. 
51 of 1,238,358 searches reached the budget.

\subsection{Maze-Unique}
\label[appendix]{app:maze-unique}
Maze-Unique \citep{eqr2026} (\texttt{locuslab/EqR-data}, subset \texttt{maze-30x30-unique-1k}) has 1,000 training and 1,000 test mazes of $30\times30$ cells, whose answers are a unique path from start to goal.
Walls are clamped off, the start and goal on, and PL of the binary path labels is applied at every open cell.
We chose the number of updates and the learning rate on 100 held-out training mazes: the fewest updates (10, 20, 50 or 100), then the smallest rate (0.3, 1 or 3), at which every run solves at least 99 of them with focused Gibbs search when trained on nested subsets of 10 to 900 mazes.
The final fits use all 1,000 mazes.

Searching a maze starts with every open cell on.
Each move sets a uniformly random cell that is not at its lowest-energy value to that value, and the search stops at a fixed point (at most 290 moves were used).
From this start, each move turns off one dead-end cell, so the search performs dead-end filling.
This choice of start matters since the empty board has lower energy than the solution on every development maze, and explains why exact solvers like toulbar2 can't find this solution.

\subsection{N-Queens}
\label[appendix]{app:nqueens}
N-Queens uses GRAM's data \citep{gram2026} without augmentation.
An input is a solution with $k$ queens removed ($k\in\{5,6,7\}$ on $8\times8$, $\{7,8,9\}$ on $10\times10$), and an input with $S$ solutions occurs $S$ times, once with each solution as its label (7,224 training and 1,240 test rows on $8\times8$; 108,234 and 18,466 on $10\times10$).
Accuracy counts the test rows whose answer is a valid board that keeps every given queen.
Cov@20 sums, over distinct test inputs, the distinct valid solutions among 20 samples, capped at $\min(S,20)$, and divides by the sum of $\min(S,20)$. 
This is the rule of EBR's evaluator, not GRAM's.
Training follows \cref{tab:recipes}, with the gradient of each $J_d$ divided by $m_d/\bar m$, where $m_d$ is the number of cell pairs at offset $d$ and $\bar m$ its mean. Without this preconditioner, the long diagonals (2 pairs, against 162 for the shortest) barely train since there is insufficient data to learn long correlations.

Each chain starts from the given queens.
A move picks a uniformly random free cell whose current state has the higher field (any free cell if there is none) and draws its state with probability proportional to $\exp(-g_i(a)/u)$.
A chain stops when every free cell has a margin of at least $\delta-10^{-3}$, or after 20,000 moves with its lowest-energy board; $N{=}1$ is the first chain, and $N{=}20$ the lowest-energy board of 20 chains.
A rule fixed before testing chose this recipe on $10\times10$ validation inputs; we report it at both sizes, a choice made after seeing the test results (on $8\times8$, the rule's own choice scores 100.00\% and 98.3\% Cov@20, against 99.95\% and 97.9\%).
To save GPU kernels, the fit builds the pair-cost matrix from the offset costs as one product with a fixed 0/1 matrix instead of by indexing.
We timed this fit and tested the models of an indexing fit, whose parameters agree with it to within $10^{-7}$ of the largest parameter.

\subsection{Graph coloring}
\label[appendix]{app:graph-coloring}
Graph coloring uses GRAM's graphs with chromatic number 3, each labeled with one 3-coloring in canonical order.
We score each graph once.
Conflict counts the edges whose two vertices share a color, summed over the test graphs.
Cov@20 is the number of distinct proper colorings in canonical order that 20 samples find, over the number of such colorings, summed over graphs (GRAM's definition).
For each pair of vertices, the $3\times3$ cost table is an affine function of all the adjacency tokens of the graph, so \CRAM{} is not told which edge belongs to which pair (7,332 and 18,660 parameters; 869 and 475 are nonzero after training).
Training follows \cref{tab:recipes}.

Each chain starts from a uniformly random coloring and makes the focused moves of Sudoku-Extreme, with focus threshold $0.45\bar h$.
It stops at the first coloring on which every pair term and input cost is at most its highest value on a training coloring.

\subsection{Hand-set energies}
\begin{figure}[ht]
\centering
\includegraphics[width=\linewidth]{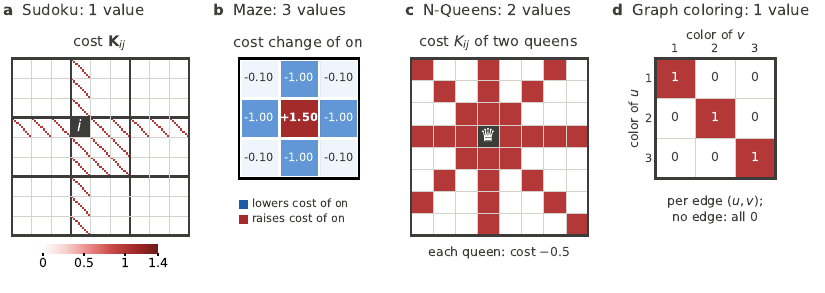}
\caption{\textbf{Hand-set weights on \CRAM{} energy.}
Each title gives the number of distinct nonzero values.
(a)~Sudoku: board cell $j$ shows the $9\times9$ cost block $\vect{\paircost}_{ij}$ between the digits of the chosen cell $i$ (rows) and of cell $j$ (columns), on the color scale of the learned costs in \cref{fig:compute}b. It is the identity on the 20 cells that share a row, column or box with $i$.
(b)~Maze-Unique: change in energy when an open cell turns on; the center is the change with no neighbor on, and each other entry is the change that a neighbor on at that offset adds.
(c)~N-Queens ($8\times8$): cost of a queen at each cell together with a queen at the dark cell, 1 when they attack each other.
(d)~Graph coloring: cost of the colors of the two vertices of an edge; pairs without an edge cost 0.}
\label{fig:hand-set}
\end{figure}

\label[appendix]{app:hand-set}
\cref{tab:main,tab:nq-gc} include \CRAM{} energies whose weights we set by hand from the rules of each task (\cref{fig:hand-set}). 
Reported numbers use the same search as the learned energy on the task.

\subsection{Looped-model baselines}
\label[appendix]{app:deviations}
We retrain every baseline on one H200 with its authors' code and recipe: HRM \citep{hrm2025} (commit \texttt{ac15626}), TRM \citep{trm2025} (\texttt{c011037}), FRM \citep{frm2026} (\texttt{6149a01}), EqR \citep{eqr2026} (\texttt{134c754}), GRAM \citep{gram2026} (\texttt{f2e32b9}) and EBR \citep{ebr2026} (\texttt{8c5271c}).
The training code of TRMP \citep{trmp2026} is not released, so we reimplement two of its recipes: the adaptive computation time (ACT) model of its Appendix B and Table 9 (row TRMP), its most accurate on Sudoku-Extreme, and the 16k-parameter factor-graph lattice model of its abstract (its Table 4), which the \S{} footnote of \cref{tab:main} reports.
The lattice model trains for 20,000 updates of a 512-board pool, as in that table, and its fit time in the footnote is the median of seven complete fits that follow a first fit in the same process, as for \CRAM{}; the first fit takes 14m46s.
GRAM's paper reports TRM and HRM on N-Queens and graph coloring without a recipe, so we train both in GRAM's code, which contains both architectures, with GRAM's per-task settings.
Every baseline trains in bf16 with torch.compile, and is scored at its final weights (their exponential moving average where the authors use one), or at a checkpoint chosen on development data (HRM, FRM and TRMP).
\CRAM{} fits, in contrast, run as eager PyTorch in single precision, without compilation or CUDA graphs.
We change a recipe only where our tests require it:
\begin{itemize}
\item \textbf{Maze-Unique recipes.} HRM, TRM, GRAM and EBR publish no Maze-Unique basline, so we run their Maze-Hard recipes unchanged. GRAM runs the TRM recipe with GRAM's settings for its non-Sudoku tasks, at global batch 384 (the largest that fits on one H200).
\item \textbf{Compilation and optimizer.} We compile the whole model where the authors' code does not (FRM; EqR on Maze-Unique), which makes training 1.8--2.2 times faster. TRM and EqR use a pure-PyTorch AdamATan2 in place of the authors' CUDA package, which adds less than 1\% to an update. TRMP trains in bf16, where its paper uses single precision for the Sudoku model.
\item \textbf{Schedules.} HRM trains on Sudoku-Extreme for 40,000 epochs, the schedule of its released checkpoint (though its README states 20,000).
\item \textbf{Test subset.} At 128 restarts, EqR is scored on the first 10,000 Sudoku-Extreme test puzzles, because the full set would take about 105 GPU-hours per seed.
\item \textbf{Published accuracies.} Our GRAM runs solve 93.61 $\pm$ 1.21\% of Sudoku-Extreme, 2.5 points below the published 96.1\%, so \cref{tab:main} shows the published value. On Maze-Unique, EqR's repository configuration (hidden size 44) solves 100.0\%, 99.6\% and 0\% with three seeds, so \cref{tab:main} shows the published 93.0\% with the parameters, FLOPs and fit time of the paper's configuration (hidden size 128). On $10\times10$ N-Queens, our TRM and HRM runs solve 6.90\% and 22.72\%, against the published 17.5\% and 37.8\%, so \cref{tab:nq-gc} shows the published values.
\end{itemize}

\subsection{Data efficiency}
\label[appendix]{app:data-efficiency}
In \cref{fig:difficulty-data}b, \CRAM{}, TRM, EBR and FRM train on the same data samples and only the number $k$ of unique training puzzles changes.
\begin{itemize}
\item \textbf{Puzzles.} For $k<1{,}000$, seed $s$ takes the first $k$ puzzles of a seed-$s$ permutation of TRM's 1,000-puzzle draw, so the subsets of one seed are nested. $k=1{,}000$ is the whole draw, and $k=3{,}000$ to 100,000 add puzzles from the official training split with the same builder.
\item \textbf{Augmentation.} Each puzzle is augmented with 99 symmetry variants (digit relabeling, transposition, and permutations of bands, stacks, rows and columns). No method adds its own augmentation.
\item \textbf{\CRAM{}}: Training is deterministic, so at $k=1{,}000$ the six seeds give one model and differ only in the search.
\item \textbf{TRM}: 65,100 updates of batch 768 at every $k$ (65,117 at $k=30{,}000$).
\item \textbf{EBR}: 60,000 updates of batch 768, scored as in \cref{tab:main}. Its loader needs at least 768 puzzle groups, so for $k<768$ each group is repeated $\lceil768/k\rceil$ times.
\item \textbf{FRM}: Stage A and Stage B of 100,000 updates each at batch 128, checkpoint chosen on 256 development puzzles.
\item \textbf{Scoring}: six runs (seeds 0--5) per point, on the 10,000 test puzzles of \cref{fig:compute}a.
\end{itemize}

\FloatBarrier
\section{Training methods for \CRAM{}}
\label[appendix]{app:alt-training}
As discussed in the paper, there are many ways to train the energy using approximate likelihood methods as per \cref{eq:energy-training}.
\Cref{tab:alt-training} compares twelve such methods across three classes (\cref{tab:methods}).
This section describes pseudolikelihood \citep{besag1975} and characterizes the twelve methods on Sudoku-Extreme and Maze-Unique.

\subsection{Pseudolikelihood, maximum likelihood, and cross-entropy}
\label[appendix]{app:pl-ce}
Pseudolikelihood (PL) replaces the model term of \cref{eq:energy-training} with a comparison between the observed solution $\truth{\vect x}$ and its single-cell changes.
Let $\mathcal F$ be the cells whose answers the input does not give, and let $\state{\vect x}^-_{i\leftarrow a}$ be $\truth{\vect x}$ with cell $i$ set to value $a$.
We minimize
\begin{equation}
\mathcal J_{\mathrm{PL}}(\vect{\weights};\truth{\vect x})
=\term{E_{\vect{\weights}}(\truth{\vect x})}{\shortstack{lower energy\\on data}}
+\term{\frac1{|\mathcal F|}\sum_{i\in\mathcal F}\log\sum_a\exp\!\left[-E_{\vect{\weights}}(\state{\vect x}^-_{i\leftarrow a})\right]}{raise energy on model ``samples''},
\label{eq:pl}
\end{equation}
where $a$ runs over all permitted values, the labeled one included.
Where a task needs it, we add local terms that depend only on the weights and the labeled solution, such as an L1 penalty on the costs.
The L1 penalty sets a cost exactly to zero where the examples show no constraint between two cells, so the fitted energy is sparse.
Below, we write $\labels$ for $\truth{\vect x}$ and $\labels^{i\leftarrow a}$ for $\state{\vect x}^-_{i\leftarrow a}$.

At unit temperature, the energy defines a joint distribution $P_{\vect{\weights}}(\belief\mid\puzzle)=\exp[-E_{\vect{\weights}}(\belief)]/Z_{\vect{\weights}}(\puzzle)$ over the set $\mathcal S(\puzzle)$ of one-hot answer arrays that respect the givens, and maximum likelihood minimizes $-\log P_{\vect{\weights}}(\labels\mid\puzzle)=E_{\vect{\weights}}(\labels)+\log Z_{\vect{\weights}}(\puzzle)$ \citep{ackley1985,lecun2006energy}.
The partition function $Z_{\vect{\weights}}(\puzzle)=\sum_{\belief'\in\mathcal S(\puzzle)}\exp[-E_{\vect{\weights}}(\belief')]$ sums over up to $C^{|\mathcal F|}$ assignments, so this objective is intractable for the tasks of this paper.

Conditioning on every labeled cell except $i$ cancels the partition function.
With $\vect e_a$ the one-hot vector of value $a$, the exact conditional of cell $i$ is a softmax,
\begin{equation}
q_i(a)
\equiv P_{\vect{\weights}}(\belief_i=\vect e_a\mid\belief_{-i}=\labels_{-i},\puzzle)
=\frac{\exp z_{ia}}{\sum_b\exp z_{ib}},
\qquad z_{ia}=-E_{\vect{\weights}}(\labels^{i\leftarrow a}),
\label{eq:pl-softmax}
\end{equation}
over the permitted values.
\Cref{eq:pl} is the mean over free cells of the cross-entropy $\ell_i=-\sum_a\truth{y}_{ia}\log q_i(a)$ of the label $\labels_i=(\truth{y}_{ia})_a$, with the familiar gradient $\partial\ell_i/\partial z_{ia}=q_i(a)-\truth{y}_{ia}$.
Every energy term without cell $i$ cancels in $q_i$, so the logits need only the pair and input costs of cell $i$, and no inference.
Each conditional is exact; the approximation is replacing the joint likelihood with a sum of conditionals \citep{besag1975,hyvarinen2006}.

\subsection{Comparing twelve objectives}
\label[appendix]{app:alt-methods}
\cref{tab:methods} lists different training objectives and how it relates to \cref{eq:energy-training}.
A relaxation runs $n$ Euler steps of \cref{eq:flow} (step size 0.5) from uniform beliefs, with clues clamped.
Each sampling method lowers the energy of the labeled solution and raises the energy of a negative, a board or belief state drawn from the model; no gradient flows through the drawing of the negative.
To keep the energies bounded, each also penalizes their squares, as in \citet{du2019}.

A few choices are specific to \CRAM{}.
Masked pseudolikelihood sets a hidden cell's state to zero, so that it adds nothing to the other cells' fields.
Denoising score matching uses the clean logits $3(\truth{\vect y}_i-\mathbf 1/C)$.
Contrastive divergence draws its moves at temperature 1, and persistent CD relabels each stored Sudoku chain together with its puzzle.
Equilibrium propagation uses $\beta=0.01$, at which its gradient has cosine similarity 0.99999 with the full-\BPTT{} gradient ($\beta=0.5$ diverged).
Implicit differentiation runs $n$ forward and $n$ adjoint iterations, without a convergence test, and full \BPTT{} uses $T=10$, selected on both tasks.

\begin{table}[ht]
\centering
\caption{\textbf{Local training objective overview.} \protect\AltClass{L}local objectives, evaluated at the labeled solution. \protect\AltClass{S}sampling methods, whose negatives get no gradient. \protect\AltClass{D}methods that differentiate through a relaxation or its fixed point. $k$, $n$, $T$, $q$ and $\sigma$ are tuned (\cref{app:alt-training}).}
\label{tab:methods}
\small
\begingroup
\setlength{\tabcolsep}{3.9pt}
\renewcommand{\arraystretch}{1.18}
\begin{tabular*}{\linewidth}{@{\extracolsep{\fill}}ll@{}}
\toprule
\textbf{\TableLabel{Training method}} & \textbf{\TableLabel{Compares the labeled solution with}}\\
\midrule
\AltClass{L}Pseudolikelihood~{\color{black!50}\hypersetup{citecolor=black!50}\citep{besag1975}} & every single-cell change (\cref{eq:pl})\\
\AltClass{L}Masked pseudolikelihood~{\color{black!50}\hypersetup{citecolor=black!50}\citep{epll2025}} & its single-cell changes, with each cell hidden with probability $q$\\
\AltClass{L}Min Probability Flow~{\color{black!50}\hypersetup{citecolor=black!50}\citep{sohldickstein2011}} & every single-cell change, weighted by $\exp(-\Delta E/2)$\\
\AltClass{L}Denoising Score Matching~{\color{black!50}\hypersetup{citecolor=black!50}\citep{vincent2011}} & its own logits plus Gaussian noise of scale $\sigma$\\
\addlinespace[2pt]
\AltClass{S}Contrastive divergence~{\color{black!50}\hypersetup{citecolor=black!50}\citep{hinton2002}} & the board after $k$ heat-bath moves from it\\
\AltClass{S}Persistent CD~{\color{black!50}\hypersetup{citecolor=black!50}\citep{tieleman2008}} & a stored chain per training board, advanced $k$ moves per visit\\
\AltClass{S}Contrastive Hebbian~{\color{black!50}\hypersetup{citecolor=black!50}\citep{movellan1991}} & the state after a relaxation of $n$ steps\\
\addlinespace[2pt]
\AltClass{D}Eq Prop~{\color{black!50}\hypersetup{citecolor=black!50}\citep{scellier2017}} & two relaxations nudged toward and away from it\\
\AltClass{D}Implicit Diff.~{\color{black!50}\hypersetup{citecolor=black!50}\citep{deq2019}} & the relaxed state, through its adjoint\\
\AltClass{D}Truncated BPTT ($t{=}1$)~{\color{black!50}\hypersetup{citecolor=black!50}\citep{hrm2025}} & the relaxed state, through the last of $T$ steps\\
\AltClass{D}Truncated BPTT ($t{=}5$)~{\color{black!50}\hypersetup{citecolor=black!50}\citep{werbos1990}} & the relaxed state, through the last 5 of $T$ steps\\
\AltClass{D}BPTT (full)~{\color{black!50}\hypersetup{citecolor=black!50}\citep{werbos1990}} & the relaxed state, through all $T$ steps\\
\bottomrule
\end{tabular*}
\endgroup
\end{table}

Every method trains the energy of \cref{tab:main} from the same weight initialization, with SGD (momentum 0.5, cosine schedule) and the L1 penalty as a soft-threshold of $\eta_k\lambda$ after each step.
All twelve methods run in one shared training harness, so their times compare with each other but \textbf{not} with the fit times of \cref{tab:main}.
The harness is slower than \cref{tab:main} because it synchronizes the GPU before and after every update, to time each update and place the development probes.

Each method tunes on seed 0, using development puzzles not in the train or test set:
\begin{itemize}
\item the learning rate: 6, 60 or 540 (Sudoku-Extreme); 0.1, 0.3 or 1 (Maze-Unique);
\item the L1 strength: 1, 0.1 or 0.01 times that of \cref{tab:main} (Sudoku-Extreme); 0, 0.003 or 0.03 (Maze-Unique);
\item its own setting: $q\in\{0.1,0.3\}$ (masked PL); $\sigma\in\{0.5,1.5\}$ (DSM); $k\in\{81,324\}$ moves on Sudoku-Extreme and $\{256,1024\}$ on Maze-Unique (CD, persistent CD); $n\in\{10,30\}$ on Sudoku-Extreme and $\{30,100\}$ on Maze-Unique (contrastive Hebbian learning, equilibrium propagation, implicit differentiation); $T\in\{10,30\}$ (truncated and full \BPTT{}); and the weight of the \citet{du2019} regularizer, 0 or 1 (class S).
\end{itemize}
Pseudolikelihood keeps the learning rate and number of updates of \cref{tab:main} and tunes only its L1 strength; every other method runs 2,000 updates (1,500 for class S on Maze-Unique).
The rule takes the highest development solve rate, then the shortest fit time.
The settings selected for each method are in the code release.

\Cref{fig:alt-training} trains each of its five methods 20 times; its x-axis is the fit time without the one-time start-up, and full \BPTT{} runs 60,000 updates.
Four differentiating methods were still improving after 2,000 updates on Sudoku-Extreme, so we gave them longer schedules (footnote of \cref{tab:alt-training}).
Full \BPTT{}, equilibrium propagation and implicit differentiation stop at incorrect boards and fail to train.
On Maze-Unique, every method solves every test maze.

Notes on \cref{tab:alt-training}:
\begin{itemize}
\item T$_{99}$ is the median over the three runs of the fit time at the first probe where their mean reaches 99\%, minus each run's one-time start-up (0.10--1.83\,s on Sudoku-Extreme and 0.64--1.22\,s on Maze-Unique), plus the time to fit the stopping threshold and focus statistic (24--75\,ms on Sudoku-Extreme; the Maze-Unique search fits none).
\item Contrastive divergence runs 324 heat-bath moves per update on Sudoku-Extreme and 256 on Maze-Unique.
\item The pseudolikelihood row is the recipe of \cref{tab:main} under this protocol. Its test solve rates equal those of \cref{tab:main}, but its fit times and FLOPs differ: it tests the checkpoint chosen on the probe (updates 16, 23 and 16 of 32 on Sudoku-Extreme; 6 of 20 on Maze-Unique), tunes the L1 strength (Maze-Unique selects 0.003, where \cref{tab:main} uses none), trains Maze-Unique on 900 mazes, and tests Sudoku-Extreme on 50,000 puzzles.
\end{itemize}

\begin{table}[!ht]
\centering
\caption{\textbf{Nine of the twelve training methods reach at least 99\% test accuracy on both tasks, and a local objective is the first to reach 99\% on the development probe of each task.}
\TableCaptionLabel{T$_{99}$}: training time until the mean development solve rate of three runs first reaches 99\% (dash: never). \TableCaptionLabel{FLOPs$_{99}$}: training arithmetic to that point. \TableCaptionLabel{Solved}: exact test solve rate (\%), mean $\pm$ SD over three runs. Classes, protocol and notes in \cref{app:alt-training}; bold: best value of each column. Every method runs in one shared training harness that is slower than the recipes of \cref{tab:main}, so compare these times with each other, not with \cref{tab:main}: a full pseudolikelihood fit on Sudoku-Extreme takes about 70\,ms here, against 42\,ms there.
}
\label{tab:alt-training}
\footnotesize
\begingroup
\setlength{\tabcolsep}{3.9pt}
\renewcommand{\arraystretch}{1.18}
\begin{tabular}{@{}l rrr rrr@{}}
\toprule
 & \multicolumn{3}{c}{\textbf{\TableLabel{Sudoku-Extreme}}} & \multicolumn{3}{c}{\textbf{\TableLabel{Maze-Unique}}}\\
\cmidrule(lr){2-4}\cmidrule(l){5-7}
\textbf{\TableLabel{Training method}} & \TableHead{T$_{99}$} & \TableHead{FLOPs$_{99}$} & \TableHead{Solved} & \TableHead{T$_{99}$} & \TableHead{FLOPs$_{99}$} & \TableHead{Solved}\\
\midrule
\AltClass{L}Pseudolikelihood~{\color{black!50}\hypersetup{citecolor=black!50}\citep{besag1975}}
  & \TableBest{\TableTime{50}{ms}} & \TableBest{\TableSci{3.42}{10}} & 100.00{\scriptsize$\,\pm$0.00} & \TableTime{43}{ms} & \TableSci{6.15}{7} & \TableBest{100.00{\scriptsize$\,\pm$0.00}}\\
\AltClass{L}Masked pseudolikelihood~{\color{black!50}\hypersetup{citecolor=black!50}\citep{epll2025}}
  & \TableTime{198}{ms} & \TableSci{1.95}{11} & 99.99{\scriptsize$\,\pm$0.00} & \TableBest{\TableTime{22}{ms}} & \TableBest{\TableSci{3.11}{7}} & \TableBest{100.00{\scriptsize$\,\pm$0.00}}\\
\AltClass{L}Min Probability Flow~{\color{black!50}\hypersetup{citecolor=black!50}\citep{sohldickstein2011}}
  & \TableTime{225}{ms} & \TableSci{1.95}{11} & 99.99{\scriptsize$\,\pm$0.00} & \TableTime{30}{ms} & \TableSci{4.22}{7} & \TableBest{100.00{\scriptsize$\,\pm$0.00}}\\
\AltClass{L}Denoising Score Matching~{\color{black!50}\hypersetup{citecolor=black!50}\citep{vincent2011}}
  & \TableTime{1.6}{s} & \TableSci{2.20}{12} & 99.98{\scriptsize$\,\pm$0.01} & \TableTime{23}{ms} & \TableSci{3.44}{7} & \TableBest{100.00{\scriptsize$\,\pm$0.00}}\\
\addlinespace[2pt]
\AltClass{S}Contrastive divergence~{\color{black!50}\hypersetup{citecolor=black!50}\citep{hinton2002}}
  & \TableTime{33.7}{s} & \TableSci{1.26}{14} & 99.99{\scriptsize$\,\pm$0.00} & \TableTime{258}{ms} & \TableSci{3.37}{9} & \TableBest{100.00{\scriptsize$\,\pm$0.00}}\\
\AltClass{S}Persistent CD~{\color{black!50}\hypersetup{citecolor=black!50}\citep{tieleman2008}}
  & \TableTime{30.2}{s} & \TableSci{1.26}{14} & 99.96{\scriptsize$\,\pm$0.06} & \TableTime{262}{ms} & \TableSci{3.37}{9} & \TableBest{100.00{\scriptsize$\,\pm$0.00}}\\
\AltClass{S}Contrastive Hebbian~{\color{black!50}\hypersetup{citecolor=black!50}\citep{movellan1991}}
  & \TableTime{348}{ms} & \TableSci{1.92}{12} & \TableBest{100.00{\scriptsize$\,\pm$0.00}} & \TableTime{111}{ms} & \TableSci{2.15}{9} & \TableBest{100.00{\scriptsize$\,\pm$0.00}}\\
\addlinespace[2pt]
\AltClass{D}Eq Prop~{\color{black!50}\hypersetup{citecolor=black!50}\citep{scellier2017}}
  & --- & --- & 16.05{\scriptsize$\,\pm$8.75}$^{\dagger}$ & \TableTime{3.7}{s} & \TableSci{6.66}{10} & \TableBest{100.00{\scriptsize$\,\pm$0.00}}\\
\AltClass{D}Implicit Diff.~{\color{black!50}\hypersetup{citecolor=black!50}\citep{deq2019}}
  & --- & --- & 0.45{\scriptsize$\,\pm$0.50} & \TableTime{2.3}{s} & \TableSci{4.23}{10} & \TableBest{100.00{\scriptsize$\,\pm$0.00}}\\
\AltClass{D}Truncated BPTT ($t{=}1$)~{\color{black!50}\hypersetup{citecolor=black!50}\citep{hrm2025}}
  & \TableTime{4.6}{s} & \TableSci{2.35}{13} & 99.57{\scriptsize$\,\pm$0.29} & \TableTime{1.1}{s} & \TableSci{7.41}{9} & \TableBest{100.00{\scriptsize$\,\pm$0.00}}\\
\AltClass{D}Truncated BPTT ($t{=}5$)~{\color{black!50}\hypersetup{citecolor=black!50}\citep{werbos1990}}
  & \TableTime{2.9}{min} & \TableSci{1.37}{15} & 99.89{\scriptsize$\,\pm$0.07}$^{\dagger}$ & \TableTime{4.0}{s} & \TableSci{1.24}{10} & \TableBest{100.00{\scriptsize$\,\pm$0.00}}\\
\AltClass{D}BPTT (full)~{\color{black!50}\hypersetup{citecolor=black!50}\citep{werbos1990}}
  & --- & --- & 48.82{\scriptsize$\,\pm$3.73}$^{\dagger}$ & \TableTime{7.7}{s} & \TableSci{1.86}{10} & \TableBest{100.00{\scriptsize$\,\pm$0.00}}\\
\bottomrule
\end{tabular}
\endgroup
\par\smallskip
{\scriptsize\color{black!55}\raggedright $^{\dagger}$Longer cosine schedules on Sudoku-Extreme, without the one-hour cap, with the learning rate chosen on the seed-0 probe.
Truncated BPTT ($t{=}5$): 60,000 updates at learning rate 6 (5.3--5.4\,min per run; 98.51\,$\pm$\,2.15\% with 2,000 updates);
Eq Prop: 240,000 updates at learning rate 6 (49--50\,min per run; 1.10\,$\pm$\,0.72\% with 2,000 updates);
BPTT (full): 3,840,000 updates at learning rate 2 (4.8--5.0\,h per run; 0.72\,$\pm$\,0.92\% with 2,000 updates, and 46.24\,$\pm$\,2.86\% after 480,000 updates).
Implicit differentiation keeps its 2,000-update cells; with 60,000 updates it solved at most 0.6\% of development puzzles.\par}
\end{table}
  
\FloatBarrier
\section{Energy descent}
\label[appendix]{app:dynamics}
\label[appendix]{app:descent}
Here we show how deterministic dynamics of \cref{eq:flow} never increase the energy.
Define the convex logsumexp function as $\mathcal L(\logits_i)=\beta^{-1}\log\sum_a\exp(\beta\state{v}_{ia})$.
Its derivative is the softmax belief, $\belief_i=\partial\mathcal L/\partial\logits_i$, and
\begin{equation}
\logits_i^{\top}\belief_i-\mathcal L(\logits_i)=\beta^{-1}\sum_a\state{p}_{ia}\log\state{p}_{ia}.
\end{equation}
so the negative-entropy term of \cref{eq:core} is the Legendre transform of $\mathcal L$.
For masked cells, all sums and derivatives run over permitted values only.
Inside the simplex, the derivative of the energy with respect to a belief is
\begin{equation}
\frac{\partial E}{\partial\belief_i}=\beta^{-1}(\log\belief_i+1)+\sum_j\vect{\symcost}_{ij}\belief_j+\vect{\inputcost}_i
=\logits_i+\sum_j\vect{\symcost}_{ij}\belief_j+\vect{\inputcost}_i+c_i,
\end{equation}
where the scalar $c_i=\beta^{-1}-\mathcal L(\logits_i)$ is added to every entry.
So \cref{eq:flow} follows $-\partial E/\partial\belief_i$ up to a constant across values, which leaves the softmax beliefs unchanged.
$\mathcal L$ is a convex function, so its Hessian is positive semidefinite:
\begin{equation}
H_i=\frac{\partial^2\mathcal L}{\partial\logits_i\,\partial\logits_i^{\top}}=\beta\big[\operatorname{diag}(\belief_i)-\belief_i\belief_i^{\top}\big]\succeq0.
\end{equation}
which is singular along the all-ones direction, because adding a constant to all logits leaves the softmax unchanged.
Since $\partial\belief_i/\partial\logits_i=H_i$, and $H_i$ removes the constant $c_i$, the chain rule and \cref{eq:flow} give
\begin{equation}
\frac{dE}{dt}=\sum_{i\in\mathcal F}\Big(\frac{\partial E}{\partial\belief_i}\Big)^{\!\top}\frac{\partial\belief_i}{\partial\logits_i}\frac{d\logits_i}{dt}
=-\sum_{i\in\mathcal F}\tau_i\dot{\logits}_i^{\top}H_i\dot{\logits}_i\le0.
\end{equation}
This holds for fixed inputs, parameters, temperature and clamps, positive time constants, and continuous time.
It does not show convergence to a unique minimum, rule out saddles, or say that a minimum is a correct solution.
Discrete steps and finite precision need their own analysis.

A board is a single-cell local minimum if no permitted single-cell change lowers its energy.
If every labeled cell strictly prefers its label given the other labels, the labeled board is such a minimum.
A small average pseudolikelihood loss does not guarantee this for every cell, and it does not say which minimum a search reaches from a puzzle.
Focused Gibbs search can pass through higher-energy boards, and it can stop at the energy threshold before it reaches a local minimum.
 
\FloatBarrier
\section{Extending \CRAM{} to Maze-Hard}
\label[appendix]{sec:maze-hard-latents}

\begin{center}
\setlength{\fboxsep}{7pt}%
\fcolorbox{black!25}{black!4}{\parbox{\dimexpr\linewidth-2\fboxsep-2\fboxrule\relax}{%
\textbf{Summary.} The Maze-Unique energy, extended with 11 learned parameters and one hidden layer that settles in its own relaxation, solves 99.9\% of the Maze-Hard test mazes after \textbf{less than 30 seconds of training} on one GPU.}}
\end{center}

On Maze-Unique, the learned energy fills dead ends from the all-open board until only the unique solution path is left (\cref{app:maze-unique}).
Maze-Hard mazes, on the other hand, contain loops that break dead-end filling, because the choice between two sides of a loop depends on distances to the goal.
With nowhere in \cref{eq:core} to store these distance calculations, \CRAM{} solves none of the Maze-Hard test mazes.

To solve this, we can add one layer of hidden neurons $\state{\vect z}$ whose state falls by a fixed amount with each step away from the goal, and let the pair costs of \cref{eq:core} read it.
Training is the pseudolikelihood of \cref{eq:pl}, as for the other tasks.

\begin{table}[!t]
\centering
\caption{\textbf{\CRAM{}+$z$ solves 99.9\% of the Maze-Hard test mazes with 29 parameters.}
Exact match to the labeled path on the 1,000 test mazes.
Baselines: accuracy published by their authors.
\CRAM{}+$z$: mean and standard deviation over three training runs (999 of 1,000 in each).}
\label{tab:maze-hard}
\begingroup
\definecolor{CRAMRow}{HTML}{EEE9F8}
\renewcommand{\TableFamilies}[2]{}
\setlength{\tabcolsep}{4pt}
\renewcommand{\arraystretch}{1.18}
\begin{tabularx}{\linewidth}{Xrrr}
\toprule
\textbf{\TableLabel{Model}} & \TableHead{Solved} & \TableHead{Params} & \TableHead{FLOPs/puzl}\\
\midrule
\TablePublished{\TableHRM{}}
  & \TableScore{74.50}{} & \TableSci{2.70}{7} & \TableSci{3.01}{12}\\
\TablePublished{\TableTRM{}}
  & \TableScore{85.30}{} & \TableSci{6.82}{6} & \TableSci{3.84}{12}\\
\TablePublished{\TableFPRM{}}
  & \TableScore{87.00}{} & \TableSci{6.83}{6} & \TableSci{4.78}{14}\\
\TablePublished{\TableNCA{}}
  & \TableScore{89.20}{} & \TableSci{4.10}{4} & \TableSci{2.00}{12}\\
\TablePublished{\TableEBR{}}
  & \TableScore{88.30}{} & \TableSci{1.77}{6} & \TableSci{1.93}{13}\\
\rowcolor{CRAMRow}
\TableCRAM{+$z$}
  & \TableBest{\TableScore{99.90}{0.00}} & \TableBest{\TableSci{2.90}{1}} & \TableBest{\TableSci{1.41}{7}}\\
\bottomrule
\end{tabularx}
\par\smallskip
{\scriptsize\color{black!55}\raggedright
$\dagger$ Accuracy from the original paper; parameter counts and FLOPs computed locally.\par
FPRM: its published setting (at most 35,000 iterations per batch of 768 mazes, restart seed 4), which solved 86.9\% in our run of the released checkpoint.\par
NCA: trained and scored on the official A* targets. EBR: lowest energy of 5 sampled trajectories, its published setting.\par
}
\endgroup
 \end{table}

\paragraph{Data.}
Maze-Hard \citep{hrm2025} has 1,000 training and 1,000 test mazes of $30\times30$ cells, with tokens $\given{x}_i\in\{\text{wall},\text{open},\text{S},\text{G}\}$ for the start S and the goal G.
The label marks one shortest path from S to G, the one that breadth-first search finds with the neighbour order down, left, right, up \citep{ebr2026}. 
Every development training maze has at least 512 shortest paths, so this tie-break decides most labels.

\begin{figure}[t]
\centering
\includegraphics[width=\linewidth]{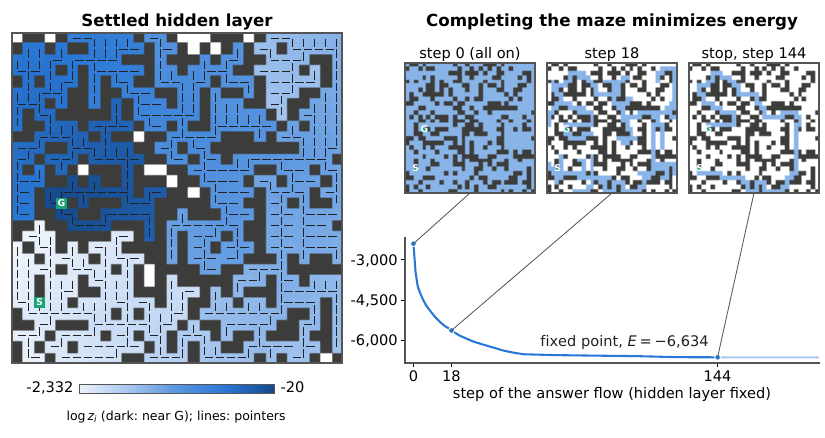}
\caption{\textbf{\CRAM{}+$z$ solves a Maze-Hard test maze in two relaxations: the hidden layer settles to a distance-like field, and the answers prune the tree of pointers down to the path.}
Left: the hidden layer on test maze 0 (seed-0 model), settled first by \cref{eq:hidden-update}: $\log\state{z}^\star_i$ (white: not connected to G; dark gray: walls) and each cell's pointer, a line toward its largest weight in \cref{eq:hidden-attention}; the pointers form a tree rooted at G.
Top right: rounded answers of the answer flow, run second with the hidden layer fixed, from the all-on board: leaves switch off first; at the stop the board equals the labeled path.
Bottom right: the answer energy (\cref{eq:core}, entropy included) decreases at every step of that flow; the lighter line continues the flow for 48 steps after the stop, to its fixed point.}
\label{fig:maze-hard}
\end{figure}

\paragraph{The hidden layer.}
Each cell is either a wall, empty, the start S or the goal G. 
A cell is ``open'' if it is not a wall.
Let $\mathcal N(i)$ be the set of open side neighbours of an open cell $i$.
Each open cell has one hidden neuron with output $\state{z}_i$ and the identity activation; $\state{z}_{\mathrm G}$ is the output of the neuron at G.
The hidden energy is
\begin{equation}
E_z(\state{\vect z})
=\sum_i\Big[\frac{\coupling{m}}2\state{z}_i^2-\frac12\sum_{j\in\mathcal N(i)}\state{z}_i\state{z}_j\Big]-\state{z}_{\mathrm G},
\label{eq:hidden-energy}
\end{equation}
with one learned mass $\coupling{m}$.
The maze enters only through which cells are open, in $\mathcal N(i)$, and through the position of G.
The first term pulls every output toward zero, the second rewards neighbouring outputs that are both large, and the third is a unit input to the neuron at G.
A cell has at most four neighbors, requiring that $\coupling{m}>4$, (enforced by learning $\log(\coupling{m}-4)$).
In each row of the Hessian $\nabla^2_{\state{\vect z}}E_z$, the diagonal entry $\coupling{m}$ is larger than the sum of the other entries, which are at most four entries of size 1, so the Hessian is positive definite.
$E_z$ is then strictly convex and has one minimum $\state{\vect z}^\star$, positive on the cells connected to G and zero elsewhere.

The neurons follow the gradient of the energy,
\begin{equation}
\tau\frac{d\state{z}_i}{dt}=-\frac{\partial E_z}{\partial\state{z}_i}=
\begin{cases}
\displaystyle-\coupling{m}\state{z}_i+\sum_{j\in\mathcal N(i)}\state{z}_j+1, & i=\mathrm G,\\[14pt]
\displaystyle-\coupling{m}\state{z}_i+\sum_{j\in\mathcal N(i)}\state{z}_j, & \text{otherwise}.
\end{cases}
\label{eq:hidden-flow}
\end{equation}
The time constant $\tau$ does not affect the fixed point, only the speed of these dynamics. 
Letting $\tau=\coupling{m}$, an Euler step of size 1 replaces every output, all at once, by the sum of its neighbours' outputs (plus 1 at G) divided by $\coupling{m}$:
\begin{equation}
\state{z}_i\leftarrow
\begin{cases}
\displaystyle\frac1{\coupling{m}}\Big(\sum_{j\in\mathcal N(i)}\state{z}_j+1\Big), & i=\mathrm G,\\[14pt]
\displaystyle\frac1{\coupling{m}}\sum_{j\in\mathcal N(i)}\state{z}_j, & \text{otherwise}.
\end{cases}
\label{eq:hidden-update}
\end{equation}
So each neuron takes the sum of its neighbours' outputs, plus the input at G, damped by the factor $1/\coupling{m}$.
This update converges to $\state{\vect z}^\star$ from any start.
Let the error of a neuron be $\state{z}_i-\state{z}^\star_i$.
One update makes the error of each neuron equal to the sum of its neighbours' errors, divided by $\coupling{m}$.
A neuron has at most four neighbours, so each update multiplies the largest error by at most $4/\coupling{m}$, which is less than 1.
From $\state{\vect z}=\vect 0$, a cell at distance $d$ from G first becomes nonzero after $d+1$ updates, so the layer settles in about as many updates as the longest distance (at most 152 on the development mazes).

\paragraph{Solving a maze.}
A maze is solved with two relaxations, one after the other (\cref{fig:maze-hard}).
First, the hidden layer settles, where \cref{eq:hidden-update} updates initial state $\state{\vect z}=\vect 0$ to fixed point.
Then, the answers start with every open cell on and follow \cref{eq:flow}.
The hidden layer is linked to the visible layer via $\coupling{L}_{ij}(\state{\vect z})$ through attention weights $\alpha_{ij}$ over $\mathcal M(i)=\mathcal N(i)\cup\{i\}$:
\begin{equation}
\begin{aligned}
\alpha_{ij}(\state{\vect z})&=\frac{e^{\coupling{\beta}_{j-i}}\,\state{z}_j}{\sum_{k\in\mathcal M(i)}e^{\coupling{\beta}_{k-i}}\,\state{z}_k},\qquad
\coupling{L}_{ij}(\state{\vect z})=\coupling{\kappa}_{j-i}\,\alpha_{ij}(\state{\vect z})+\coupling{\kappa}_{i-j}\,\alpha_{ji}(\state{\vect z}).
\end{aligned}
\label{eq:hidden-attention}
\end{equation}
Each answer update of \cref{eq:euler} then adds $\eta\sum_{j\in\mathcal M(i)}\coupling{L}_{ij}(\state{\vect z})\,\state{p}_j$ to the Maze-Unique update of cell $i$, and learnable $\coupling{\beta}_\delta$ (one per direct neighbor, and itself) allows the model to learn Maze-Hard's preference for direction.
The answers $\state{p}_i\in[0,1]$ have the same Maze-Unique energy as in \cref{app:maze-unique}, such that their update is:

\begin{equation}
\begin{aligned}
\logits_i&\leftarrow(1-\eta)\,\logits_i
-\eta\Big(\sum_j\big(\term{\vect{\symcost}_{ij}}{Maze-Unique}-\term{\coupling{L}_{ij}(\state{\vect z})\operatorname{diag}(0,1)}{links to the hidden layer}\big)\belief_j+\vect{\inputcost}_i\Big),\\
\belief_i&=\softmax(\logits_i).
\end{aligned}
\label{eq:hidden-answer-update}
\end{equation}
Here $\logits_i\in\R^2$ holds the logits of the off and on states of cell $i$, and $\vect{\symcost}_{ij}$ and $\vect{\inputcost}_i$ are the $2\times2$ pair costs and the input cost of the Maze-Unique energy, so with $\coupling{L}=0$ this is \cref{eq:euler} for Maze-Unique; $\operatorname{diag}(0,1)$ keeps only the on--on entry, so a link lowers the cost of two linked cells being on together.
The links raise the on logit of cell $i$ by the on probabilities of itself and its neighbours, weighted by their links.
Walls stay off, and S and G stay on.

The fixed point of this optimization is the predicted path.

\textbf{Training} uses pseudolikelihood training as in Maze-Unique, and an adjoint solver for the hidden parameters, without differentiating through either relaxation.
The model has 29 total learned parameters: the 18 of the Maze-Unique energy, five $\coupling{\kappa}$, five $\coupling{\beta}$ and $\coupling{m}$: with $\coupling{\kappa}=0$ it is the Maze-Unique model.

\paragraph{Results.}
On the development mazes, each of three training runs solves 100 of 100.
Trained on all 1,000 training mazes and scored once on the test set, each of three runs solves 999 of 1,000, with one incorrect stop and no exhausted budget (\cref{tab:maze-hard}).
All three fail on the same maze, where a dead end of six cells along the maze's border, next to the path, stays on: its cells support each other through the $3\times3$ pair costs, so the flow does not prune it.
On the development mazes, learning an input per token in place of the input at G solves 0 of 300, and one gate per pair in place of the normalized window weights solves 0 of 100.
A fit takes about half a minute on one H200, with the hidden-layer relaxation and its adjoint each run as one fused Triton kernel per maze. \FloatBarrier
 \ifdefined\IncludeChecklist
\clearpage
\section*{NeurIPS Paper Checklist}

\begin{enumerate}

\item {\bf Claims}
    \item[] Question: Do the main claims made in the abstract and introduction accurately reflect the paper's contributions and scope?
    \item[] Answer: \answerTODO{} %
    \item[] Justification: \justificationTODO{}
    \item[] Guidelines:
    \begin{itemize}
        \item The answer \answerNA{} means that the abstract and introduction do not include the claims made in the paper.
        \item The abstract and/or introduction should clearly state the claims made, including the contributions made in the paper and important assumptions and limitations. A \answerNo{} or \answerNA{} answer to this question will not be perceived well by the reviewers. 
        \item The claims made should match theoretical and experimental results, and reflect how much the results can be expected to generalize to other settings. 
        \item It is fine to include aspirational goals as motivation as long as it is clear that these goals are not attained by the paper. 
    \end{itemize}

\item {\bf Limitations}
    \item[] Question: Does the paper discuss the limitations of the work performed by the authors?
    \item[] Answer: \answerTODO{} %
    \item[] Justification: \justificationTODO{}
    \item[] Guidelines:
    \begin{itemize}
        \item The answer \answerNA{} means that the paper has no limitation while the answer \answerNo{} means that the paper has limitations, but those are not discussed in the paper. 
        \item The authors are encouraged to create a separate ``Limitations'' section in their paper.
        \item The paper should point out any strong assumptions and how robust the results are to violations of these assumptions (e.g., independence assumptions, noiseless settings, model well-specification, asymptotic approximations only holding locally). The authors should reflect on how these assumptions might be violated in practice and what the implications would be.
        \item The authors should reflect on the scope of the claims made, e.g., if the approach was only tested on a few datasets or with a few runs. In general, empirical results often depend on implicit assumptions, which should be articulated.
        \item The authors should reflect on the factors that influence the performance of the approach. For example, a facial recognition algorithm may perform poorly when image resolution is low or images are taken in low lighting. Or a speech-to-text system might not be used reliably to provide closed captions for online lectures because it fails to handle technical jargon.
        \item The authors should discuss the computational efficiency of the proposed algorithms and how they scale with dataset size.
        \item If applicable, the authors should discuss possible limitations of their approach to address problems of privacy and fairness.
        \item While the authors might fear that complete honesty about limitations might be used by reviewers as grounds for rejection, a worse outcome might be that reviewers discover limitations that aren't acknowledged in the paper. The authors should use their best judgment and recognize that individual actions in favor of transparency play an important role in developing norms that preserve the integrity of the community. Reviewers will be specifically instructed to not penalize honesty concerning limitations.
    \end{itemize}

\item {\bf Theory assumptions and proofs}
    \item[] Question: For each theoretical result, does the paper provide the full set of assumptions and a complete (and correct) proof?
    \item[] Answer: \answerTODO{} %
    \item[] Justification: \justificationTODO{}
    \item[] Guidelines:
    \begin{itemize}
        \item The answer \answerNA{} means that the paper does not include theoretical results. 
        \item All the theorems, formulas, and proofs in the paper should be numbered and cross-referenced.
        \item All assumptions should be clearly stated or referenced in the statement of any theorems.
        \item The proofs can either appear in the main paper or the supplemental material, but if they appear in the supplemental material, the authors are encouraged to provide a short proof sketch to provide intuition. 
        \item Inversely, any informal proof provided in the core of the paper should be complemented by formal proofs provided in appendix or supplemental material.
        \item Theorems and Lemmas that the proof relies upon should be properly referenced. 
    \end{itemize}

    \item {\bf Experimental result reproducibility}
    \item[] Question: Does the paper fully disclose all the information needed to reproduce the main experimental results of the paper to the extent that it affects the main claims and/or conclusions of the paper (regardless of whether the code and data are provided or not)?
    \item[] Answer: \answerTODO{} %
    \item[] Justification: \justificationTODO{}
    \item[] Guidelines:
    \begin{itemize}
        \item The answer \answerNA{} means that the paper does not include experiments.
        \item If the paper includes experiments, a \answerNo{} answer to this question will not be perceived well by the reviewers: Making the paper reproducible is important, regardless of whether the code and data are provided or not.
        \item If the contribution is a dataset and\slash or model, the authors should describe the steps taken to make their results reproducible or verifiable. 
        \item Depending on the contribution, reproducibility can be accomplished in various ways. For example, if the contribution is a novel architecture, describing the architecture fully might suffice, or if the contribution is a specific model and empirical evaluation, it may be necessary to either make it possible for others to replicate the model with the same dataset, or provide access to the model. In general. releasing code and data is often one good way to accomplish this, but reproducibility can also be provided via detailed instructions for how to replicate the results, access to a hosted model (e.g., in the case of a large language model), releasing of a model checkpoint, or other means that are appropriate to the research performed.
        \item While NeurIPS does not require releasing code, the conference does require all submissions to provide some reasonable avenue for reproducibility, which may depend on the nature of the contribution. For example
        \begin{enumerate}
            \item If the contribution is primarily a new algorithm, the paper should make it clear how to reproduce that algorithm.
            \item If the contribution is primarily a new model architecture, the paper should describe the architecture clearly and fully.
            \item If the contribution is a new model (e.g., a large language model), then there should either be a way to access this model for reproducing the results or a way to reproduce the model (e.g., with an open-source dataset or instructions for how to construct the dataset).
            \item We recognize that reproducibility may be tricky in some cases, in which case authors are welcome to describe the particular way they provide for reproducibility. In the case of closed-source models, it may be that access to the model is limited in some way (e.g., to registered users), but it should be possible for other researchers to have some path to reproducing or verifying the results.
        \end{enumerate}
    \end{itemize}

\item {\bf Open access to data and code}
    \item[] Question: Does the paper provide open access to the data and code, with sufficient instructions to faithfully reproduce the main experimental results, as described in supplemental material?
    \item[] Answer: \answerTODO{} %
    \item[] Justification: \justificationTODO{}
    \item[] Guidelines:
    \begin{itemize}
        \item The answer \answerNA{} means that paper does not include experiments requiring code.
        \item Please see the NeurIPS code and data submission guidelines (\url{https://neurips.cc/public/guides/CodeSubmissionPolicy}) for more details.
        \item While we encourage the release of code and data, we understand that this might not be possible, so \answerNo{} is an acceptable answer. Papers cannot be rejected simply for not including code, unless this is central to the contribution (e.g., for a new open-source benchmark).
        \item The instructions should contain the exact command and environment needed to run to reproduce the results. See the NeurIPS code and data submission guidelines (\url{https://neurips.cc/public/guides/CodeSubmissionPolicy}) for more details.
        \item The authors should provide instructions on data access and preparation, including how to access the raw data, preprocessed data, intermediate data, and generated data, etc.
        \item The authors should provide scripts to reproduce all experimental results for the new proposed method and baselines. If only a subset of experiments are reproducible, they should state which ones are omitted from the script and why.
        \item At submission time, to preserve anonymity, the authors should release anonymized versions (if applicable).
        \item Providing as much information as possible in supplemental material (appended to the paper) is recommended, but including URLs to data and code is permitted.
    \end{itemize}

\item {\bf Experimental setting/details}
    \item[] Question: Does the paper specify all the training and test details (e.g., data splits, hyperparameters, how they were chosen, type of optimizer) necessary to understand the results?
    \item[] Answer: \answerTODO{} %
    \item[] Justification: \justificationTODO{}
    \item[] Guidelines:
    \begin{itemize}
        \item The answer \answerNA{} means that the paper does not include experiments.
        \item The experimental setting should be presented in the core of the paper to a level of detail that is necessary to appreciate the results and make sense of them.
        \item The full details can be provided either with the code, in appendix, or as supplemental material.
    \end{itemize}

\item {\bf Experiment statistical significance}
    \item[] Question: Does the paper report error bars suitably and correctly defined or other appropriate information about the statistical significance of the experiments?
    \item[] Answer: \answerTODO{} %
    \item[] Justification: \justificationTODO{}
    \item[] Guidelines:
    \begin{itemize}
        \item The answer \answerNA{} means that the paper does not include experiments.
        \item The authors should answer \answerYes{} if the results are accompanied by error bars, confidence intervals, or statistical significance tests, at least for the experiments that support the main claims of the paper.
        \item The factors of variability that the error bars are capturing should be clearly stated (for example, train/test split, initialization, random drawing of some parameter, or overall run with given experimental conditions).
        \item The method for calculating the error bars should be explained (closed form formula, call to a library function, bootstrap, etc.)
        \item The assumptions made should be given (e.g., Normally distributed errors).
        \item It should be clear whether the error bar is the standard deviation or the standard error of the mean.
        \item It is OK to report 1-sigma error bars, but one should state it. The authors should preferably report a 2-sigma error bar than state that they have a 96\% CI, if the hypothesis of Normality of errors is not verified.
        \item For asymmetric distributions, the authors should be careful not to show in tables or figures symmetric error bars that would yield results that are out of range (e.g., negative error rates).
        \item If error bars are reported in tables or plots, the authors should explain in the text how they were calculated and reference the corresponding figures or tables in the text.
    \end{itemize}

\item {\bf Experiments compute resources}
    \item[] Question: For each experiment, does the paper provide sufficient information on the computer resources (type of compute workers, memory, time of execution) needed to reproduce the experiments?
    \item[] Answer: \answerTODO{} %
    \item[] Justification: \justificationTODO{}
    \item[] Guidelines:
    \begin{itemize}
        \item The answer \answerNA{} means that the paper does not include experiments.
        \item The paper should indicate the type of compute workers CPU or GPU, internal cluster, or cloud provider, including relevant memory and storage.
        \item The paper should provide the amount of compute required for each of the individual experimental runs as well as estimate the total compute. 
        \item The paper should disclose whether the full research project required more compute than the experiments reported in the paper (e.g., preliminary or failed experiments that didn't make it into the paper). 
    \end{itemize}
    
\item {\bf Code of ethics}
    \item[] Question: Does the research conducted in the paper conform, in every respect, with the NeurIPS Code of Ethics \url{https://neurips.cc/public/EthicsGuidelines}?
    \item[] Answer: \answerTODO{} %
    \item[] Justification: \justificationTODO{}
    \item[] Guidelines:
    \begin{itemize}
        \item The answer \answerNA{} means that the authors have not reviewed the NeurIPS Code of Ethics.
        \item If the authors answer \answerNo, they should explain the special circumstances that require a deviation from the Code of Ethics.
        \item The authors should make sure to preserve anonymity (e.g., if there is a special consideration due to laws or regulations in their jurisdiction).
    \end{itemize}

\item {\bf Broader impacts}
    \item[] Question: Does the paper discuss both potential positive societal impacts and negative societal impacts of the work performed?
    \item[] Answer: \answerTODO{} %
    \item[] Justification: \justificationTODO{}
    \item[] Guidelines:
    \begin{itemize}
        \item The answer \answerNA{} means that there is no societal impact of the work performed.
        \item If the authors answer \answerNA{} or \answerNo, they should explain why their work has no societal impact or why the paper does not address societal impact.
        \item Examples of negative societal impacts include potential malicious or unintended uses (e.g., disinformation, generating fake profiles, surveillance), fairness considerations (e.g., deployment of technologies that could make decisions that unfairly impact specific groups), privacy considerations, and security considerations.
        \item The conference expects that many papers will be foundational research and not tied to particular applications, let alone deployments. However, if there is a direct path to any negative applications, the authors should point it out. For example, it is legitimate to point out that an improvement in the quality of generative models could be used to generate Deepfakes for disinformation. On the other hand, it is not needed to point out that a generic algorithm for optimizing neural networks could enable people to train models that generate Deepfakes faster.
        \item The authors should consider possible harms that could arise when the technology is being used as intended and functioning correctly, harms that could arise when the technology is being used as intended but gives incorrect results, and harms following from (intentional or unintentional) misuse of the technology.
        \item If there are negative societal impacts, the authors could also discuss possible mitigation strategies (e.g., gated release of models, providing defenses in addition to attacks, mechanisms for monitoring misuse, mechanisms to monitor how a system learns from feedback over time, improving the efficiency and accessibility of ML).
    \end{itemize}
    
\item {\bf Safeguards}
    \item[] Question: Does the paper describe safeguards that have been put in place for responsible release of data or models that have a high risk for misuse (e.g., pre-trained language models, image generators, or scraped datasets)?
    \item[] Answer: \answerTODO{} %
    \item[] Justification: \justificationTODO{}
    \item[] Guidelines:
    \begin{itemize}
        \item The answer \answerNA{} means that the paper poses no such risks.
        \item Released models that have a high risk for misuse or dual-use should be released with necessary safeguards to allow for controlled use of the model, for example by requiring that users adhere to usage guidelines or restrictions to access the model or implementing safety filters. 
        \item Datasets that have been scraped from the Internet could pose safety risks. The authors should describe how they avoided releasing unsafe images.
        \item We recognize that providing effective safeguards is challenging, and many papers do not require this, but we encourage authors to take this into account and make a best faith effort.
    \end{itemize}

\item {\bf Licenses for existing assets}
    \item[] Question: Are the creators or original owners of assets (e.g., code, data, models), used in the paper, properly credited and are the license and terms of use explicitly mentioned and properly respected?
    \item[] Answer: \answerTODO{} %
    \item[] Justification: \justificationTODO{}
    \item[] Guidelines:
    \begin{itemize}
        \item The answer \answerNA{} means that the paper does not use existing assets.
        \item The authors should cite the original paper that produced the code package or dataset.
        \item The authors should state which version of the asset is used and, if possible, include a URL.
        \item The name of the license (e.g., CC-BY 4.0) should be included for each asset.
        \item For scraped data from a particular source (e.g., website), the copyright and terms of service of that source should be provided.
        \item If assets are released, the license, copyright information, and terms of use in the package should be provided. For popular datasets, \url{paperswithcode.com/datasets} has curated licenses for some datasets. Their licensing guide can help determine the license of a dataset.
        \item For existing datasets that are re-packaged, both the original license and the license of the derived asset (if it has changed) should be provided.
        \item If this information is not available online, the authors are encouraged to reach out to the asset's creators.
    \end{itemize}

\item {\bf New assets}
    \item[] Question: Are new assets introduced in the paper well documented and is the documentation provided alongside the assets?
    \item[] Answer: \answerTODO{} %
    \item[] Justification: \justificationTODO{}
    \item[] Guidelines:
    \begin{itemize}
        \item The answer \answerNA{} means that the paper does not release new assets.
        \item Researchers should communicate the details of the dataset\slash code\slash model as part of their submissions via structured templates. This includes details about training, license, limitations, etc. 
        \item The paper should discuss whether and how consent was obtained from people whose asset is used.
        \item At submission time, remember to anonymize your assets (if applicable). You can either create an anonymized URL or include an anonymized zip file.
    \end{itemize}

\item {\bf Crowdsourcing and research with human subjects}
    \item[] Question: For crowdsourcing experiments and research with human subjects, does the paper include the full text of instructions given to participants and screenshots, if applicable, as well as details about compensation (if any)? 
    \item[] Answer: \answerTODO{} %
    \item[] Justification: \justificationTODO{}
    \item[] Guidelines:
    \begin{itemize}
        \item The answer \answerNA{} means that the paper does not involve crowdsourcing nor research with human subjects.
        \item Including this information in the supplemental material is fine, but if the main contribution of the paper involves human subjects, then as much detail as possible should be included in the main paper. 
        \item According to the NeurIPS Code of Ethics, workers involved in data collection, curation, or other labor should be paid at least the minimum wage in the country of the data collector. 
    \end{itemize}

\item {\bf Institutional review board (IRB) approvals or equivalent for research with human subjects}
    \item[] Question: Does the paper describe potential risks incurred by study participants, whether such risks were disclosed to the subjects, and whether Institutional Review Board (IRB) approvals (or an equivalent approval/review based on the requirements of your country or institution) were obtained?
    \item[] Answer: \answerTODO{} %
    \item[] Justification: \justificationTODO{}
    \item[] Guidelines:
    \begin{itemize}
        \item The answer \answerNA{} means that the paper does not involve crowdsourcing nor research with human subjects.
        \item Depending on the country in which research is conducted, IRB approval (or equivalent) may be required for any human subjects research. If you obtained IRB approval, you should clearly state this in the paper. 
        \item We recognize that the procedures for this may vary significantly between institutions and locations, and we expect authors to adhere to the NeurIPS Code of Ethics and the guidelines for their institution. 
        \item For initial submissions, do not include any information that would break anonymity (if applicable), such as the institution conducting the review.
    \end{itemize}

\item {\bf Declaration of LLM usage}
    \item[] Question: Does the paper describe the usage of LLMs if it is an important, original, or non-standard component of the core methods in this research? Note that if the LLM is used only for writing, editing, or formatting purposes and does \emph{not} impact the core methodology, scientific rigor, or originality of the research, declaration is not required.
    \item[] Answer: \answerTODO{} %
    \item[] Justification: \justificationTODO{}
    \item[] Guidelines:
    \begin{itemize}
        \item The answer \answerNA{} means that the core method development in this research does not involve LLMs as any important, original, or non-standard components.
        \item Please refer to our LLM policy in the NeurIPS handbook for what should or should not be described.
    \end{itemize}

\end{enumerate} \fi
\end{document}